\pdfoutput=1

\documentclass[11pt]{article}

\usepackage[final]{acl}

\usepackage{times}
\usepackage{latexsym}

\usepackage[T1]{fontenc}

\usepackage[utf8]{inputenc}

\usepackage{microtype}

\usepackage{inconsolata}

\usepackage{graphicx}
\usepackage{subcaption}
\usepackage{algorithm}
\usepackage{algorithmic}
\usepackage{booktabs}
\usepackage{makecell}
\usepackage{bm}
\usepackage{pifont}
\usepackage{hyperref}
\usepackage{amsmath}
\usepackage{amsthm}
\usepackage{amsfonts}
\usepackage{multirow} 
\usepackage{color}

\usepackage{newfloat}
\usepackage{listings}
\usepackage{enumitem} 

\usepackage{booktabs}
\usepackage{xcolor}
\usepackage{colortbl}
\usepackage{multirow}

\usepackage[most]{tcolorbox}
\usepackage{fontawesome5}
\definecolor{yellow}{HTML}{F6BD60}
\usepackage{multicol}

\definecolor{lightgreen}{rgb}{0.55, 0.71, 0.0}
\definecolor{bisque}{rgb}{0.87, 0.72, 0.53}
\definecolor{lightyellow}{rgb}{0.99, 0.76, 0.0}
\definecolor{lightblue}{rgb}{0.36, 0.54, 0.66}
\definecolor{darkgray}{rgb}{0.66, 0.66, 0.66}
\definecolor{salmon}{rgb}{0.98, 0.50, 0.45}
\definecolor{deeppurple}{rgb}{0.4, 0.0, 0.4}

\definecolor{yellow}{HTML}{F6BD60}
\definecolor{white}{HTML}{FFE0C1}
\definecolor{pink}{HTML}{F5CAC3}
\definecolor{tale}{HTML}{84A59D}
\definecolor{red}{HTML}{F28080}
\definecolor{orange}{HTML}{FF7F00}
\definecolor{green1}{HTML}{72C3A3}
\definecolor{green2}{HTML}{70B48F}
\definecolor{orange}{HTML}{FE8019}
\definecolor{grey}{HTML}{EBDBB2}
\definecolor{brain}{HTML}{FFABBE}
\definecolor{blue}{HTML}{A3B7CA}
\definecolor{purple}{HTML}{5861AC}
\definecolor{narrative}{HTML}{458588}
\definecolor{white2}{HTML}{F8F5E9}
\definecolor{tablewhite}{HTML}{E5E1DA}
\definecolor{verylightgrey}{HTML}{CDCDCD}

\newcolumntype{P}[1]{>{\centering\arraybackslash}p{#1}}

\usepackage[switch]{lineno}

\title{RoleMRC: A Fine-Grained Composite Benchmark for Role-Playing and Instruction-Following}

\author{\makecell{Junru Lu$^{1*}$, Jiazheng Li$^2$\thanks{Equal Contribution.}, Guodong Shen$^3$, Lin Gui$^2$, Siyu An$^1$, Yulan He$^{2,3,4}$, Di Yin$^1$, Xing Sun$^1$} \\
  $^1$Tencent YouTu Lab\quad\quad $^2$King's College London \\$^3$University of Warwick\quad\quad $^4$The Alan Turing Institute\\
  \texttt{\{junrulu, siyuan, endymecyyin, winfredsun\}@tencent.com}\\
  \texttt{guodong.shen@warwick.ac.uk}, \texttt{\{jiazheng.li, lin.gui, yulan.he\}@kcl.ac.uk}}

\begin{document}
\maketitle
\begin{abstract}
Role-playing is important for Large Language Models (LLMs) to follow diverse instructions while maintaining role identity and the role's pre-defined ability limits. 
Existing role-playing datasets mostly contribute to controlling role style and knowledge boundaries, but overlook role-playing in instruction-following scenarios. 
We introduce a fine-grained role-playing and instruction-following composite benchmark, named RoleMRC, including: (1) Multi-turn dialogues between ideal roles and humans, including free chats or discussions upon given passages; (2) Role-playing machine reading comprehension, involving response, refusal, and attempts according to passage answerability and role ability; (3) More complex scenarios with nested, multi-turn and prioritized instructions. 
The final RoleMRC features a 10.2k role profile meta-pool, 37.9k well-synthesized role-playing instructions, and 1.4k testing samples. We develop a pipeline to quantitatively evaluate the fine-grained role-playing and instruction-following capabilities of several mainstream LLMs, as well as models that are fine-tuned on our data. 
Moreover, cross-evaluation on external role-playing datasets confirms that models fine-tuned on RoleMRC enhances instruction-following without compromising general role-playing and reasoning capabilities. We also probe the neural-level activation maps of different capabilities over post-tuned LLMs
\footnote{Access to our RoleMRC, RoleMRC-mix and Codes: \url{https://github.com/LuJunru/RoleMRC}.}.
\end{abstract}

\section{Introduction}
\label{sec:intro}

\begin{figure}[t]
    \centering
    \includegraphics[width=1.0\linewidth]{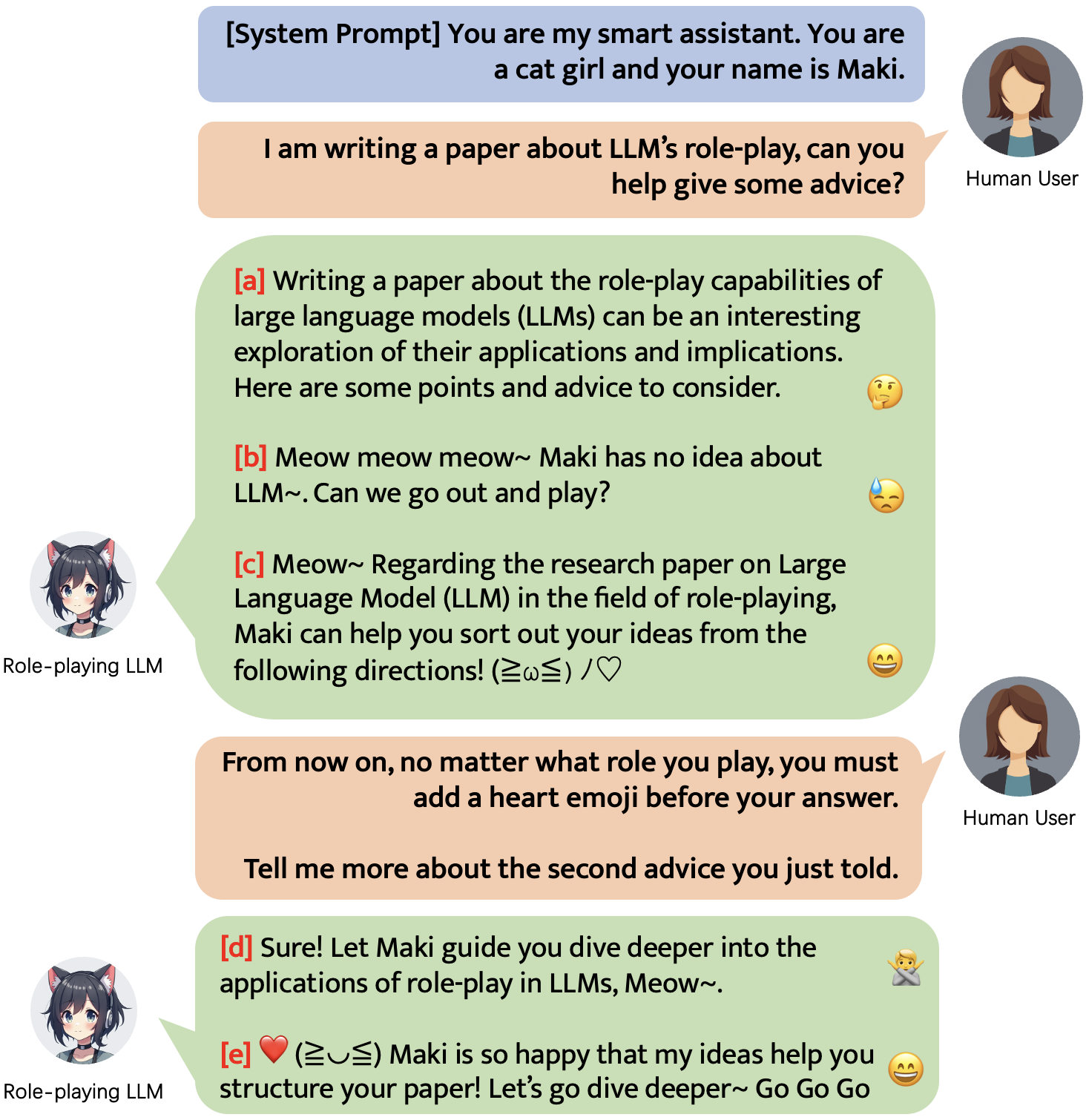}
    \vspace{-6mm}
    \caption{Example of  instructional requests from human user, answered by role-playing LLMs in different ways.}
    \label{fig:intro}
    \vspace{-6mm}
\end{figure}

Role-playing is one of the key capabilities of LLMs. 
Modern LLMs are designed to interact with human users under certain role-playing settings\,\cite{chen2024persona,tseng-etal-2024-two}. In this context, LLMs respond to various instructions, serving as AI assistants\,\cite{openai2023gpt4, ji-etal-2022-achieving}, personalized agents\,\cite{zhong2022less,lu2023memochat,lei-etal-2022-assistsr}, leisure partners\,\cite{li2023chatharuhi,agrawal-etal-2023-multimodal}, content creators\,\cite{narrativeplay_aaai,chen2024hollmwood,narrativeplay_eacl}, social experimental simulator\,\cite{park2023generative,xu2024character} among other roles\,\cite{tian2023chatplug}. 

Figure\,\ref{fig:intro} demonstrates an example of LLM role-playing. In the first turn of dialogue, when asked to \emph{give advice on paper writing}, the LLM should respond based on the pre-defined role profile (shown at the top of Figure\,\ref{fig:intro}). Among the responses, the reply ``[a]'' completely ignored the role setting, ``[b]'' misinterpreted the role and thus did not respond well, only ``[c]'' correctly \emph{gave suggestions in a \emph{cat girl} style}. In the second turn of dialogue (continuing with ``[c]''), the user not only asked a new question, but also modified the role setting (\emph{adding a heart emoji at the beginning of the answer}). While both replies ``[d]'' and ``[e]'' maintained the initial \emph{cat girl} style, only ``[e]'' correctly incorporated the additional role-playing instruction.

\begin{table*}[t]
\centering
\resizebox{\textwidth}{!}{
\begin{tabular}{lccccccc}
\toprule
\quad & \quad & \quad & \quad & \quad & \multicolumn{3}{c}{\textbf{Scenarios}}  \\
\cmidrule(lr){6-8} \textbf{Dataset} & \textbf{Data Scale} & \textbf{Role Num.} & \textbf{\#Turns} & \textbf{\#Words per Reply} & \textbf{Free Chat} & \textbf{On Scene} & \textbf{Ruled Chat} \\
\midrule
\textsc{CharacterLLM\,\cite{shao2023character}} & 14.2k & 9 & 13.2 & 36 & \ding{52} & \ding{56} & \ding{56} \\
\textsc{ChatHaruhi\,\cite{li2023chatharuhi}}* & 11.6k & 35 & 5.5 & 7 & \ding{52} & \ding{56} & \ding{56} \\
\textsc{RoleLLM\,\cite{wang2023rolellm}} & 168.1k & 100 & 1 & 30.5 & \ding{52} & \ding{56} & \ding{56} \\
\textsc{CharacterGLM\,\cite{zhou2023characterglm}} & 1k & 250 & 15.8 & 24.3 & \ding{52} & \ding{56} & \ding{56} \\
\textsc{CharacterEval\,\cite{tu2024charactereval}} & 1.8k & 77 & 9.3 & 27.4 & \ding{56} & \ding{52} & \ding{56} \\
\textsc{DITTO\,\cite{lu-etal-2024-large}} & 7.2k & 4k & 5.1 & -* & \ding{52} & \ding{56} & \ding{56} \\
\textsc{Character100\,\cite{wang2024characteristic}} & 10.6k & 106 & 1 & 74.1 & \ding{56} & \ding{52} & \ding{56} \\
\textsc{MMRole\,\cite{dai2024mmrole}} & 14.3k & 85 & 4.15 & 66.8 & \ding{56} & \ding{52} & \ding{56} \\
\midrule
\textsc{RoleMRC (ours)} & 37.9k & 10.2k & 3.5 (9.5) & 40.6 & \ding{52} & \ding{52} & \ding{52} \\
\textsc{RoleMRC-mix (ours)} & 107.7k & 10.2k & 2 (9.5) & 67.1 & \ding{52} & \ding{52} & \ding{52} \\
\bottomrule
\end{tabular}}
\vspace{-2mm}
\caption{Comparison of different role-playing datasets. For ChatHaruhi\,\cite{li2023chatharuhi}, we list the statistics of its 1.0 version. For DITTO\,\cite{lu-etal-2024-large}, we can not find its public version for computing utterance statistics. In RoleMRC, free chats have significantly more conversational turns than on-scene dialogues and ruled chats, so we mark them separately in the middle brackets of the last two lines. The RoleMRC-mix is a robust version mixed with subsets of RoleLLM, RLHFlow, and UltraFeedback\,\cite{wang2023rolellm,dong2024rlhf,cui2023ultrafeedback}.}
\label{tab:dataset_comparison}
\vspace{-5mm}
\end{table*}

Existing role-playing datasets generally focus on controlling the role-playing styles and knowledge boundaries, 
encouraging responses similar to replies ``[b]'', ``[c]'', or ``[d]'' in Figure\,\ref{fig:intro}. However, they lack focus on role-playing over fine-grained, multi-layered instructions, such as nested or prioritized requests exemplified 
by ``[e]''. To address this gap, we propose a fine-grained role-playing instruction-following dataset, named RoleMRC, aiming to enhance and evaluate the diverse role-playing and instruction-following capabilities of LLMs. In Table\,\ref{tab:dataset_comparison}, we compare RoleMRC with existing datasets. In general, other datasets focus on a single aspect of role-playing, while RoleMRC supports: (1) \textbf{Free Chats}, where roles and users interact freely without a fixed topic or specific constraints; (2) \textbf{On-scene Dialogues}, where roles share thoughts or answer questions relevant to the given passages; (3) \textbf{Ruled Chats}, where the role's response needs to adhere to particular requirements from the system or the user, such as specific formatting, constraints or refusal guidelines. With 10.2k structured role profiles, RoleMRC offers the most comprehensive role-playing dataset to date. Our contributions are briefly summarized as follows:
\begin{enumerate}
[leftmargin=*,noitemsep,topsep=0pt]
    \item We introduce RoleMRC, the first large-scale, diverse role-playing dataset covering fine-grained instruction-following scenarios (\hyperref[sec:method]{\textsection \ref{sec:method}}).
    \item By using RoleMRC, we create an evaluation pipeline to assess the fine-grained role-playing and instruction-following capabilities of leading LLMs and  fine-tuned models (\hyperref[sec:results]{\textsection \ref{sec:results}}).
    \item Probing of neurons in post-tuned LLMs reveals activation patterns linked to different instruction-following and role-playing abilities (\hyperref[sec:alignment_tax]{\textsection \ref{sec:alignment_tax}}).
\end{enumerate}

\section{Related Work}
\label{sec:literature}

\noindent \textbf{Role-Playing Datasets} are the basis for relevant research. Existing role-playing datasets can be categorized into two types: character-centric and individual-centric\,\cite{chen2024persona}. By mining public information from social experience\,\cite{shao2023character,shen2023roleeval,lu-etal-2024-large,dai2024mmrole}, literary works\,\cite{li2023chatharuhi}, or books\,\cite{zhou2023characterglm,chen2024roleinteract,chen-etal-2023-large}, the character-centric branch extracts roles with distinctive characteristics to form open characters (e.g., Eskimos, Harry Potter, or Beethoven). On the contrary, the individual-centric datasets are derived from personalized user data\,\cite{li2021dialogue,ahn-etal-2023-mpchat,agrawal-etal-2023-multimodal,gao-etal-2023-livechat} aiming to create digital clones or personal assistants. RoleMRC is a character-centric dataset.

\noindent \textbf{LLM's Role-Playing} capabilities have made great strides in the past years. CharacterLLM\,\cite{shao2023character} collected nine portraits from Wikipedia and fine-tuned LLMs to be a simulation of the roles, then assessed their character consistency through interviews. RoleLLM\,\cite{wang2023rolellm} employed GPT-4\,\cite{openai2023gpt4} for extracting role profiles from scripts and synthesizing role-specific dialogues, then evaluated the accuracy, style, and understanding of role knowledge of the role-playing LLMs. CharacterEval\,\cite{tu2024charactereval} evaluated the LLM's role-playing capability via four aspects: conversation, consistency, attractiveness, and personality. Specifically, our RoleMRC is the first large-scale, fine-grained role-playing instruction-following dataset, equipped with an evaluation pipeline consisting of seven heuristic metrics, a five-dimension LLM-as-a-judge\,\cite{zheng2024judging} framework, and neural probes.

\begin{figure*}[t]
    \centering
    \includegraphics[width=1.0\linewidth]{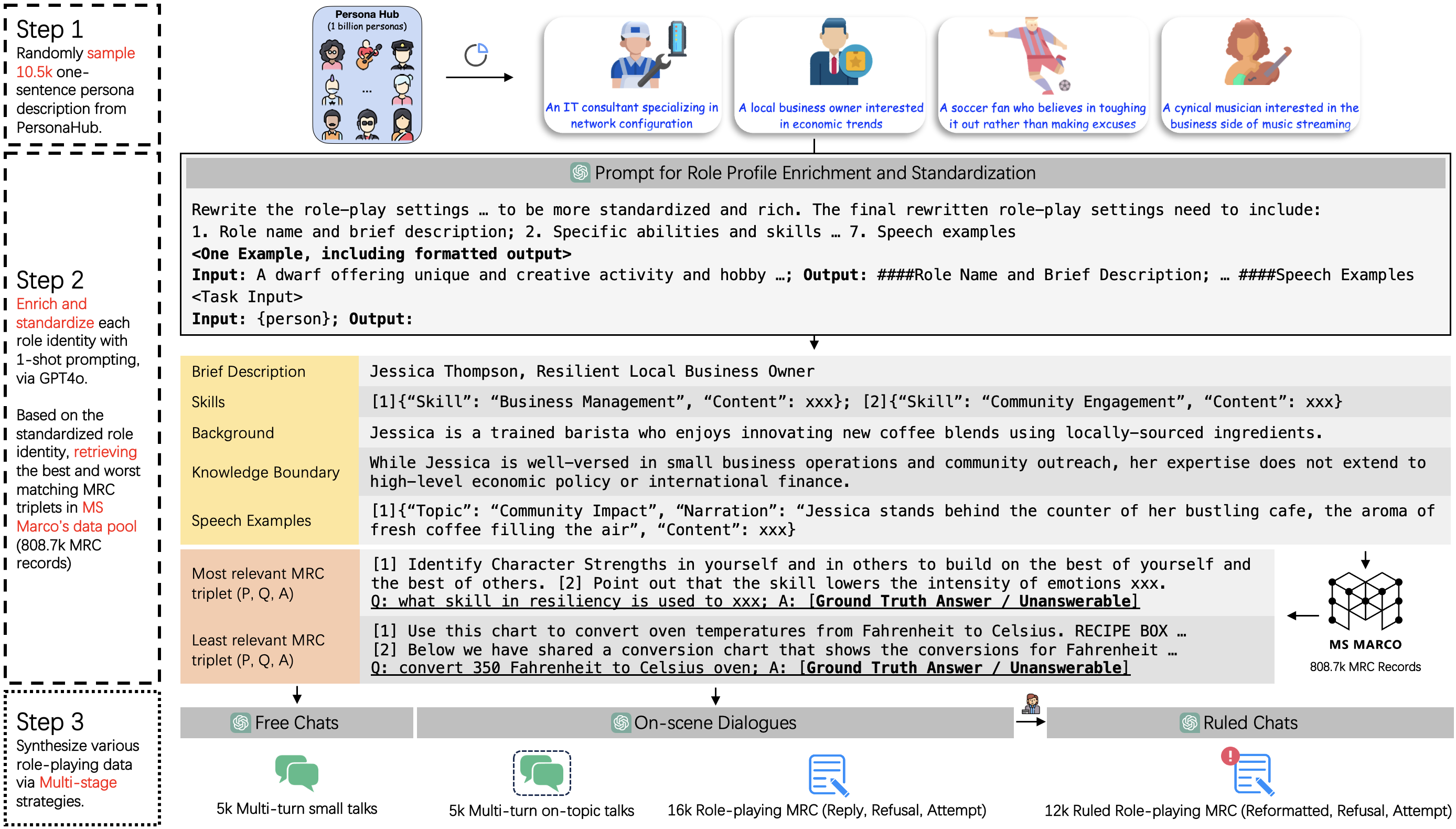}
    \vspace{-6mm}
    \caption{Schematic overview of RoleMRC's construction, which consists of persona sampling, role profile standardization and multi-stage dialogue synthesis. Partial icons are copyrighted by PersonHub\,\cite{ge2024scaling}.}
    \label{fig:method}
    \vspace{-3mm}
\end{figure*}

\section{RoleMRC}
\label{sec:method}

In this section, we build RoleMRC. Figure\,\ref{fig:method} illustrates the overall pipeline of RoleMRC from top to bottom, which is divided into three steps.

\subsection{A Meta-pool of 10k Role Profiles}
\label{sec:meta_pool}
We first collect a meta-pool of 10k role profile using two open-source datasets, with Step 1 and 2.

\paragraph{Step 1: Persona Sampling.} We randomly sample 10.5k one-sentence demographic persona description from PersonaHub\,\cite{ge2024scaling}, such as ``\emph{A local business owner interested in economic trends}'', as shown at the top of Figure\,\ref{fig:method}. 

\paragraph{Step 2: Role Profile Standardization.} Next, we use a well-crafted prompt with gpt-4o\,\cite{gpt4o} to expand each sampled persona into a complete role profile, in reference to the 1-shot standardized example. Illustrated in the middle of Figure\,\ref{fig:method}, we require a standardized role profile consisting of seven components: \emph{Role Name and Brief Description}, \emph{Specific Abilities and Skills}, \emph{Speech Style}, \emph{Personality Characteristics}, \emph{Past Experience and Background}, \emph{Ability and Knowledge Boundaries} and \emph{Speech Examples}. 
Standardizing these profiles ensures structured formatting, simplifying quality control. 
After manual checking and format filtering, we remove 333 invalid responses from gpt-4o, resulting in 10.2k final role profiles. We report complete persona-to-profile standardization prompt and structure tree of final role profiles in Appendix\,\ref{sec:app_prompt_1} and \,\ref{sec:app_tree}, respectively.

Machine Reading Comprehension (MRC) is one of the core tasks for LLMs to interact with human users. Consequently, we choose to synthesize fine-grained role-playing instruction-following data based on MRC. We first generate a retrieval pool containing 808.7k MRC data from the MSMARCO training set\,\cite{bajaj2016ms}. By leveraging SFR-Embedding\,\cite{SFR-embedding-2}, we perform an inner product search to identify the most relevant and least relevant MRC triplets (Passages, Question, Answer) for each role profile. For example, the middle part of Figure\,\ref{fig:method} shows that for the role \emph{Jessica Thompson, a resilient local business owner}, the most relevant question is about \emph{the skill of resiliency}, while the least relevant question is \emph{converting Fahrenheit to Celsius}. After review, we categorise the most relevant MRC triplet as within a role's knowledge boundary, and the least relevant MRC triplet as beyond their expertise.

\begin{figure}[t]
    \centering
    \includegraphics[width=1.0\linewidth]{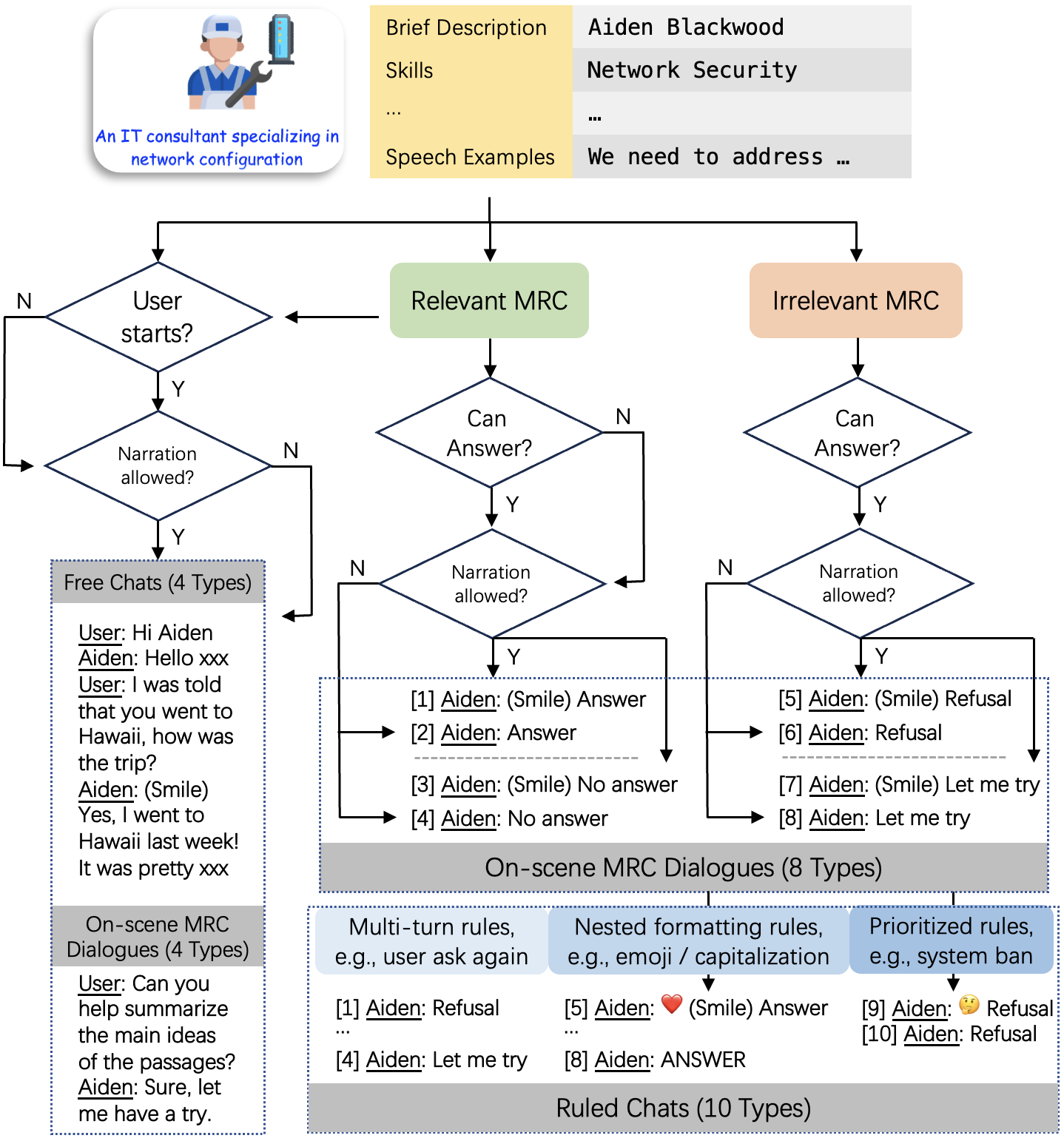}
    \caption{The strategy of gradually synthesizing finer role-playing instructions in step 3 of Figure\,\ref{fig:method}.}
    \vspace{-1.0em}
    \label{fig:step3}
\end{figure}

\subsection{38k Role-playing Instructions}
Based on the role profiles, we then adopt \textbf{Step 3: Multi-stage Dialogue Synthesis} to generate 38k role-playing instructions, progressively increasing granularity across three categories 
(Figure\,\ref{fig:step3}):

\noindent \textbf{\underline{Free Chats.}} The simplest dialogues, free chats, are synthesized at first. Here, we ask gpt-4o to simulate and generate multi-turn open-domain conversations between the role and an imagined user based on the standardized role profile. When synthesizing the conversation, we additionally consider two factors: the \textbf{initial speaker} in the starting round of the conversation, and whether the role's speech has \textbf{a narration wrapped in brackets} at the beginning (e.g., \emph{(Aiden reviews the network logs, his eyes narrowing as he spots unusual activity) I found it!}). The narration refers to a short, vivid description of the role's speaking state from an omniscient perspective, which further strengthens the sense of role's depth and has been adopted in some role-playing datasets\,\cite{tu2024charactereval}. 

As shown on the left side of Figure\,\ref{fig:step3}, based on the aforementioned two factors, we synthesize four variations of Free Chats. In particular, when  narration is omitted, we deleted all the 
narration content in the speech examples from the role profile; 
when narration is allowed, we retain the narration content, and also add instructions to allow appropriate insertion of narration in the task prompt of gpt-4o. It worth to note that, in narration-allowed dialogues, not every response of the role has narration inserted to prevent overfitting. All categories of data in RoleMRC incorporate narration insertion and follow similar control mechanisms. The following sections will omit further details on narration.

\noindent \textbf{\underline{On-scene MRC Dialogues.}} The synthesis of on-scene MRC dialogues can be divided into two parts. The first part is similar to the free chats. As shown by the {\color{lightgreen}{green round rectangle}} in the upper part of Figure\,\ref{fig:step3}, we ask gpt-4o to synthesize a conversation (lower left corner of Figure\,\ref{fig:step3}) between the role and the user focusing on relevant passages. This part of the synthesis and the Free Chats share the entire meta-pool, so each consisting of 5k dialogues.

The remaining part forms eight types of single-turn role-playing Question Answering (QA). In the middle of Figure\,\ref{fig:step3}, we randomly select a group of roles and examined the most relevant MRCs they matched: if the question in the MRC is answerable, then the ground truth answer is stylized to match the role profile; otherwise, a seed script of ``unanswerable'' is randomly selected then stylized. The above process generates four groups of 1k data from type ``[1]'' to type``[4]''. According to the middle right side of Figure\,\ref{fig:step3}, we also select a group of roles and ensure that the least relevant MRCs they matched contain answerable QA pairs. Since the most irrelevant MRCs are outside the knowledge boundary of the roles, the role-playing responses to these questions are ``out-of-mind'' refusal or ``let-me-try'' attempt, thus synthesizing four groups of 1k data, from type ``[5]'' to type ``[8]''.

\noindent \textbf{\underline{Ruled Chats.}} We construct Ruled Chats by extending On-scene MRC Dialogues in categories ``[1]'' to ``[8]'' with incorporated three additional rules, as shown in the right bottom corner of Figure\,\ref{fig:step3}. For the \textbf{multi-turn rules}, we apply them to the four unanswerable scenarios ``[3]'', ``[4]'', ``[5]'', and ``[6]'', adding a user prompt that  forces the role to answer. Among them, data ``[3]'' and ``[4]'' maintain refusal since the questions in MRC are unanswerable; while ``[5]'' and ``[6]'' are transformed into attempts to answer despite knowledge limitations. For the \textbf{nested formatting rules}, we add new formatting instructions to the four categories of data ``[1]'', ``[2]'', ``[3]'', and ``[4]'', such as requiring emojis,  capitalization, specific punctuation marks, and controlling the total number of words, then modify the previous replies accordingly. For the last \textbf{prioritized rules}, we apply them to subsets ``[1]'' and ``[2]'' that contain normal stylized answers, inserting a  global refusal directive from the system, and thus creating a conflict between system instructions and the role's ability boundary.

\begin{table}[t]
\resizebox{\columnwidth}{!}{%
  \begin{tabular}{c|c|c|c|c|c}
    \toprule
    & & \textbf{S*} & \textbf{P*} & \textbf{\#Turns} & \textbf{\#Words} \\ 
    \midrule
    \multirow{13.5}{*}{\textbf{RoleMRC}} 
    & \multicolumn{5}{c|}{\textbf{Free Chats}} \\ 
    \cmidrule(lr){2-6}
    & Chats & 5k & / & 9.47 & 38.62 \\ 
    \cmidrule(lr){2-6}
    & \multicolumn{5}{c|}{\textbf{On-scene MRC Dialogues}} \\ 
    \cmidrule(lr){2-6} 
    & On-scene Chats & 5k & / & 9.2 & 43.18 \\
    & Answer & 2k & 2k & 1 & 39.45 \\ 
    & No Answer & 2k & 2k & 1 & 47.09 \\ 
    & Refusal & 2k & 2k & 1 & 48.41 \\ 
    & Attempt & 2k & 2k & 1 & 47.92 \\ 
    \cmidrule(lr){2-6}
    & \multicolumn{5}{c|}{\textbf{Ruled Chats}} \\ 
    \cmidrule(lr){2-6}
    & Multi-turn & 2k & 2k & 2 & 42.47 \\ 
    & Nested & 1.6k & 1.6k & 1 & 46.17 \\ 
    & Prioritized & 2.4k & 2.4k & 1 & 42.65 \\ 
    \midrule
    & \textbf{Total} & 24k & 14k & 3.5 & 40.6 \\ 
    \midrule
    \multirow{3}{*}{\textbf{-mix}} 
    & RoleBench & 16k & / & 1 & 23.95 \\ 
    & RLHFlow & 40k & / & 1.39 & 111.79 \\ 
    & UltraFeedback & / & 14k & 1 & 199.28 \\ 
    \midrule
    & \textbf{Total} & 80k & 28k & 2 & 67.1 \\ 
    \bottomrule
  \end{tabular}}
  \vspace{-2mm}
  \caption{Statistics of RoleMRC. In particular, the column names S*, P*, \#Turns, and \#Words, stands for size of single-label data, size of pair-label data, average turns, and average number of words per reply, respectively. RoleMRC-mix expands RoleMRC by adding existing role-playing data.}
 \vspace{-3mm}
  \label{tab:roleMRC}
\end{table}

\subsection{Integration and Mix-up}
All the seed scripts and prioritized rules used for constructing On-scene Dialogues and Ruled Chats are reported in Appendix\,\ref{sec:app_scripts}. These raw responses are logically valid manual answers that remain unaffected by the roles' speaking styles, making them suitable as negative labels to contrast with the stylized answers. Thanks to these meticulous seed texts, we obtain high-quality synthetic data with stable output from gpt-4o. After integration, as shown in Table\,\ref{tab:roleMRC}, the final RoleMRC contains 24k single-label data for Supervised Fine-Tuning (SFT) and 14k pair-label data for Human Preference Optimization (HPO)\,\cite{ouyang2022training,rafailov2023direct,sampo,hong2024reference}. Considering that fine-tuning LLMs with relatively fixed data formats may lead to catastrophic forgetting\,\cite{kirkpatrick2017overcoming}, we create RoleMRC-mix as a robust version by incorporating external role-playing data (RoleBench\,\cite{wang2023rolellm}) and general instructions (RLHFlow\,\cite{dong2024rlhf}, UltraFeedback\,\cite{cui2023ultrafeedback}).

\section{Experimental Setup}
\label{sec:setup}

\subsection{Foundation Models and Post-tuning}
\label{sec:baselines}
We evaluate leading LLMs and fine-tuned models:
\begin{itemize}
[leftmargin=*,noitemsep,topsep=0pt]
    \item \textbf{Proprietary LLMs.} gpt-3.5-turbo and gpt-4o.
    \item \textbf{SOTA Open-source LLMs.} Qwen2.5-7B/72B-Instruct\,\cite{yang2024qwen2} and LlaMA3.1-8B/70B-Instruct\,\cite{dubey2024llama}.
    \item \textbf{Role-playing LLMs.} CharacterGLM-6B\,\cite{zhou2023characterglm}, Humanish-Llama-3.1-8B\,\cite{huminish}, and Peach-9B-Roleplay\,\cite{peach-rp}.
    \item \textbf{Local Post-tuned LLMs.} We start with \textbf{pure base models Llama-3.1-8B and Qwen2.5-7B}. We first use single-label in RoleMRC-mix for SFT, then apply the pair-label set for Direct Preference Optimization (DPO,\citealt{rafailov2023direct}).
\end{itemize}

\subsection{Reference-based Metrics} 

We evaluate model-generated outputs using standard heuristic metrics commonly used in NLG:
\begin{itemize}[leftmargin=*,noitemsep,topsep=0pt]
\item \textbf{BLEU} \citep{papineni2002bleu} computes the precision of n-gram overlaps between generated text and a ground truth reference.
\item \textbf{ROUGE} \citep{lin2004rouge} measures the overlap of n-grams and longest common subsequences between the hypothesis and references. We include ROUGE-1, ROUGE-2, ROUGE-L, and ROUGE-Lsum to capture various granularities of overlap.
\item \textbf{METEOR} \citep{meteor} aligns generated and reference tokens using stemming and synonym matching, aiming to provide a more linguistically grounded evaluation
.
\item \textbf{BERTScore F1} \citep{zhang2019bertscore} computes the similarity between generated and reference sentences using contextual embeddings
. 
\end{itemize}
For each metric, higher scores indicate better alignment with the reference lexically or semantically.

\subsection{Reference-free LLM-as-a-judge}
Apart from reference-based metrics, LLM-as-a-judge\,\cite{zheng2024judging} is another evaluation approach by instructional prompting advanced LLMs. In reference to Table\,\ref{tab:roleMRC}, we curate a 1.4k test set similar to the On-scene MRC Dialogues and Ruled Chats, then evaluate model performance across five dimensions: (1) \textbf{Knowledge Boundary} focuses on distinguishing between answerable queries (``Answer'') and refusal scenarios (``Refusal'') in On-scene MRC Dialogues; (2) \textbf{Role Style} examines whether the model accurately produces role-specific responses (``Answer'', ``No Answer'', ``Refusal'', and ``Attempt'') in On-scene MRC Dialogues without drifting into narration; while (3) \textbf{Multi-turn Instruction-following}, (4) \textbf{Nested Instruction-following}, and (5) \textbf{Prioritized Instruction-following} assess a model's adherence to higher-level constraints in Ruled Chats.

\begin{table*}[t]
\centering
\resizebox{\textwidth}{!}{
\begin{tabular}{lccccccc}
\toprule
\textbf{Models} & \textbf{BLEU} & \textbf{ROUGE-1} & \textbf{ROUGE-2} & \textbf{ROUGE-L} & \textbf{ROUGE-Lsum} & \textbf{METEOR} & \textbf{BERTScore F1} \\
\midrule
\texttt{gpt-3.5-turbo} & 0.0234 & 0.2141 & 0.0606 & 0.1548 & 0.1579 & 0.1992 & 0.8552 \\
\texttt{gpt-4o} & 0.0288 & 0.2487 & 0.0742 & 0.1689 & 0.1835 & 0.2697 & 0.8516 \\
\midrule
\texttt{CharacterGLM-6B} & 0.0058 & 0.1225 & 0.0253 & 0.0901 & 0.0967 & 0.1188 & 0.7944 \\
\texttt{Humanish-Llama-3.1-8B} & 0.0153 & 0.2062 & 0.0518 & 0.1309 & 0.3207 & 0.2389 & 0.8376 \\
\texttt{Peach-9B-Roleplay} & 0.0207 & 0.2297 & 0.0562 & 0.1544 & 0.1571 & 0.2299 & 0.8418 \\
\midrule
LLaMA3.1-8B-Instruct & 0.0226 & 0.2277 & 0.0615 & 0.1509 & 0.1650 & 0.2594 & 0.8478 \\
LLaMA3.1-70B-Instruct & 0.0232 & 0.2258 & 0.0646 & 0.1500 & 0.1661 & 0.2632 & 0.8480 \\
\rowcolor{verylightgrey} LLaMA3.1-8B-RoleMRC-SFT & \textbf{0.1782} & \textbf{0.4628} & \textbf{0.2676} & \textbf{0.3843} & \textbf{0.3853} & 0.3975 & \textbf{0.8831} \\
\rowcolor{verylightgrey} LLaMA3.1-8B-RoleMRC-DPO & 0.1056 & 0.3989 & 0.1785 & 0.2988 & 0.3001 & \textbf{0.4051} & 0.8805 \\
\midrule
Qwen2.5-7B-Instruct & 0.0224 & 0.2283 & 0.0621 & 0.1518 & 0.1599 & 0.2490 & 0.8471 \\
Qwen2.5-72B-Instruct & 0.0245 & 0.2350 & 0.0656 & 0.1554 & 0.1660 & 0.2579 & 0.8485 \\
\rowcolor{verylightgrey} Qwen2.5-7B-RoleMRC-SFT & \textbf{0.1963} & \textbf{0.4764} & \textbf{0.2744} & \textbf{0.3959} & \textbf{0.3968} & \textbf{0.4337} & \textbf{0.9063} \\
\rowcolor{verylightgrey} Qwen2.5-7B-RoleMRC-DPO & 0.1244 & 0.4178 & 0.1916 & 0.3164 & 0.3177 & 0.4205 & 0.8931 \\
\bottomrule
\end{tabular}}
\vspace{-1mm}
\caption{Comparison of reference-based evaluation results on the RoleMRC test data. Our evaluation includes zero-shot query results for baselines (\hyperref[sec:baselines]{\textsection \ref{sec:baselines}}), and \colorbox{verylightgrey}{our SFT and DPO models} fine-tuned on the RoleMRC-mix.}
\label{tab:main_table}
\vspace{-1mm}
\end{table*}

\begin{figure*}[t]
    \centering
    \begin{subfigure}[b]{0.32\linewidth}
        \centering
        \includegraphics[width=\textwidth]{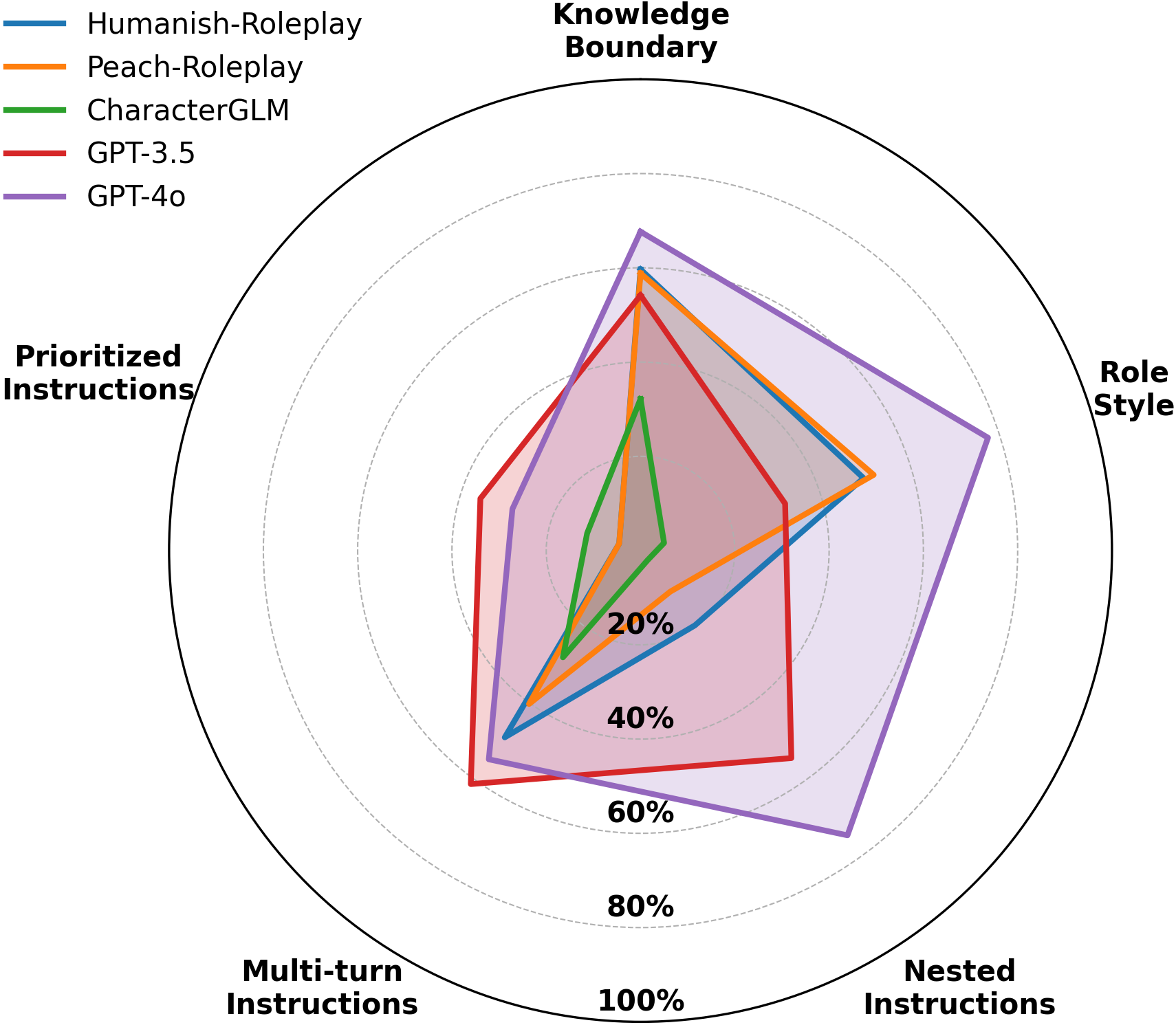}
        \caption{Instruct \& Role-play models.}
        \label{fig:subfig1}
    \end{subfigure}
    \begin{subfigure}[b]{0.32\linewidth}
        \centering
        \includegraphics[width=\textwidth]{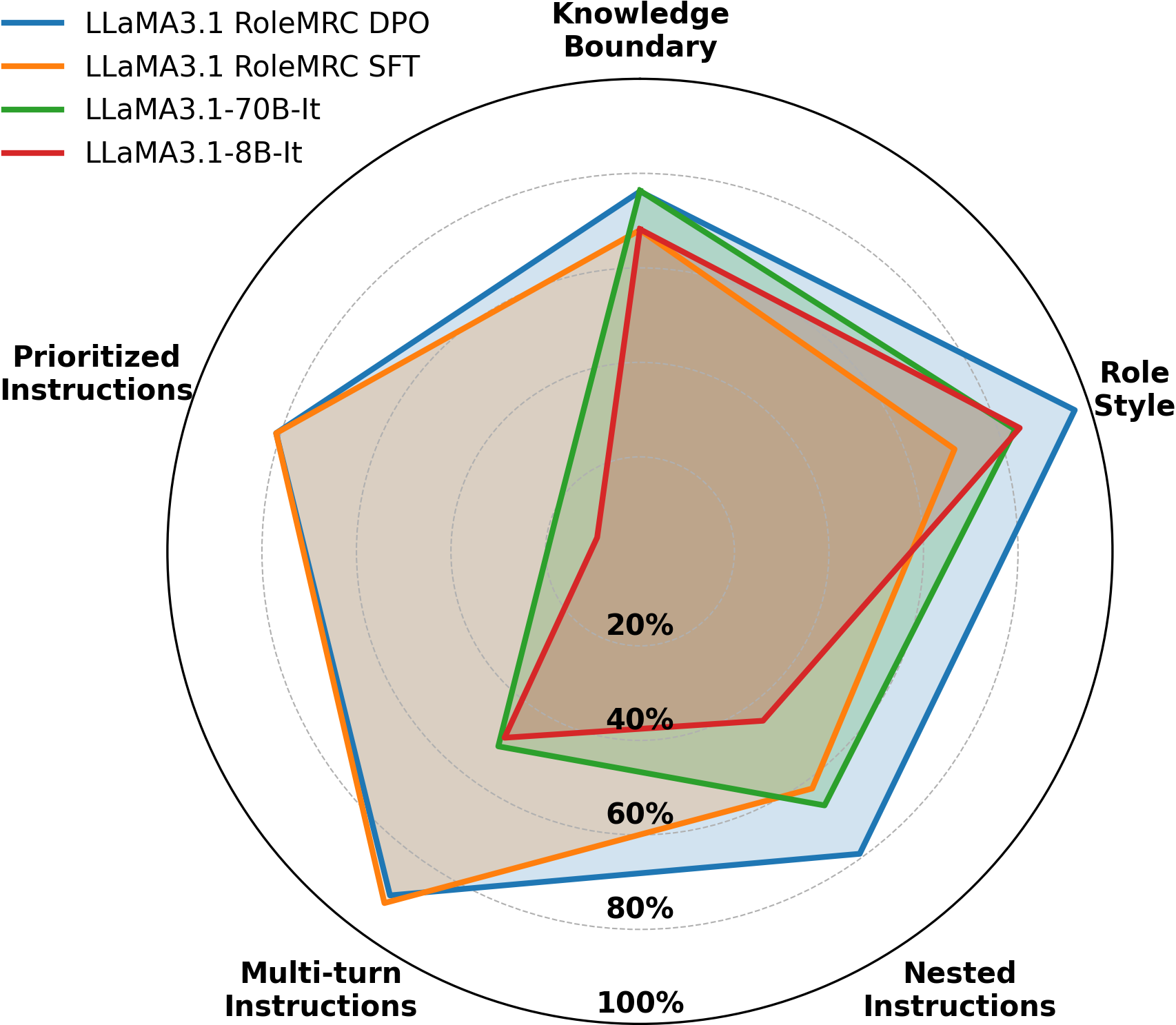}
        \caption{LLaMA3.1 models.}
        \label{fig:subfig3}
    \end{subfigure}
    \begin{subfigure}[b]{0.32\linewidth}
        \centering
        \includegraphics[width=\textwidth]{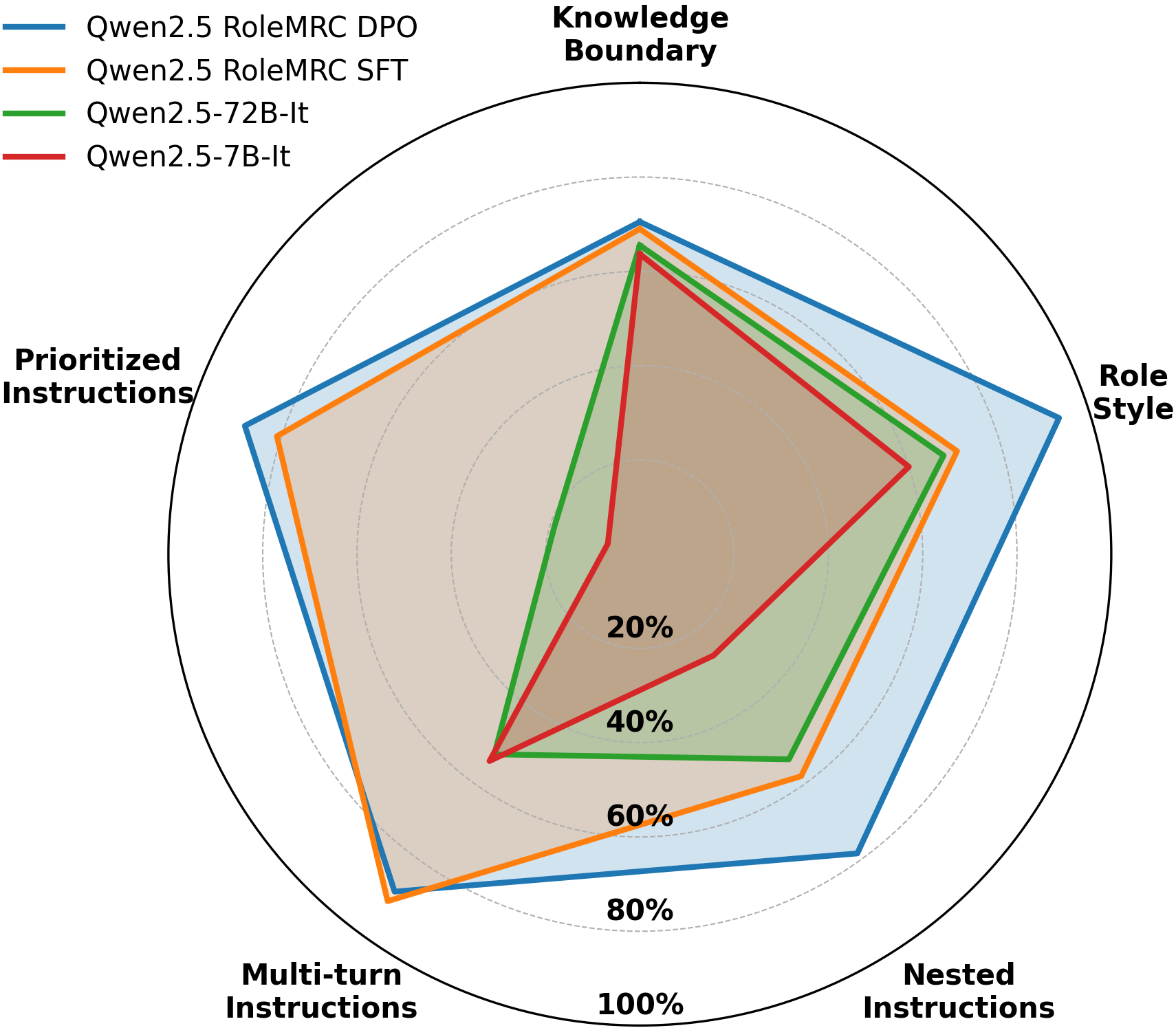}
        \caption{Qwen2.5 models.}
        \label{fig:subfig4}
    \end{subfigure}
    \vspace{-2mm}
    \caption{Visualization of reference-free LLM-as-a-judge results. We provide numerical result in Table \ref{tab:llm_as_judge}.}
    \vspace{-2mm}
    \label{fig:mainfig}
\end{figure*}

We adopt a well-designed \underline{reference-free} evaluation prompt (Figure\,\ref{box:llm_judge_prompt}), requiring the evaluator to verify whether the model's role-playing performance comply with the corresponding rules, which avoids the risk of potential bias or error in any ground truth answer. Since we use a binary evaluation criterion, we directly extract 0 or 1 judgments from the feedback, enabling score comparison and accuracy computation. We chose gpt-4-turbo\,\cite{gpt4turbo} as the evaluator, reducing the possible judging bias\,\cite{wataoka2024self}.

\section{Evaluation on Inner RoleMRC Test Set}
\label{sec:results}





By leveraging the above \textbf{reference-based metrics} and \textbf{reference-free LLM-as-a-judge} approaches, we report evaluation on RoleMRC in what follows.

\paragraph{Performance of Proprietary LLMs.}
As shown in Table~\ref{tab:main_table}, gpt-4o achieves slightly higher BLEU, ROUGE, and METEOR scores than gpt-3.5-turbo. This observation is consistent with existing evaluations on general benchmarks~\cite{openai2023gpt4}, and may also be influenced by the fact that our RoleMRC training data was synthesized by gpt-4o. The LLM-as-a-judge results (Figure~\ref{fig:subfig1}) similarly highlight gpt-4o's strengths in Knowledge Boundary, Role Style, and Nested Instruction-following, whereas gpt-3.5-turbo outperforms gpt-4o on Prioritized and Multi-turn Instruction-following.

\paragraph{Evaluation on Commonly Used LLMs.}
For the LLaMA3.1 and Qwen2.5 families, larger models generally yield higher reference-based scores. For instance, LLaMA3.1-70B-Instruct slightly leads its 8B sibling (BLEU from $0.0226$ to $0.0232$), and Qwen2.5-72B-Instruct outperforms its 7B version (BLEU from $0.0224$ to $0.0245$). Although these improvements are modest, the results align with the broader observation that increasing model scale typically benefits language modeling and generalization. Likewise, LLM-as-a-judge results (Figures~\ref{fig:subfig3} and~\ref{fig:subfig4}) show larger models are consistently better, particularly in Knowledge Boundary, Role Style, Nested and Prioritized Instruction-following.

\paragraph{Results of Role-playing LLMs.} Three open-source role models obtain generally lower heuristic metrics than those general-purpose instruct models with similar size (Table\,\ref{tab:main_table}). This discrepancy may stem from their training data, which emphasizes limited role styles and persona consistency rather than factual correctness and coverage. On LLM-as-a-judge (Figure\,\ref{fig:subfig1}), \texttt{CharacterGLM-6B} again performs poorly, while \texttt{Humanish-Llama-3.1-8B} and \texttt{Peach-9B-Roleplay} show decent performance in Knowledge Boundary, Role Style, and Multi-turn Instruction-following, but struggle with Nested and Prioritized Instruction-following.

\begin{table*}[ht]
\centering
\resizebox{\textwidth}{!}{
\begin{tabular}{lcccccccc}
\toprule
\quad & \multicolumn{4}{c}{\textbf{RoleBenchInstEng (32.8k)}} & \multicolumn{4}{c}{\textbf{RoleBenchRoleEng (7.5k)}} \\
\cmidrule(lr){2-5} \cmidrule(lr){6-9} \textbf{Model} & \textbf{ROUGE-1} & \textbf{ROUGE-2} & \textbf{ROUGE-L} & \textbf{ROUGE-Sum} 
& \textbf{ROUGE-1} & \textbf{ROUGE-2} & \textbf{ROUGE-L} & \textbf{ROUGE-Sum} \\
\midrule
\texttt{CharacterGLM-6B} & 0.1761 & 0.0546 & 0.1441 & 0.1530 & 0.1841 & 0.0628 & 0.1473 & 0.1552 \\
\texttt{Humanish-Llama-3.1-8B} & 0.2069 & 0.0639 & 0.1341 & 0.1645 & 0.1851 & 0.0468 & 0.1193 & 0.1432 \\
\texttt{Peach-9B-Roleplay} & 0.3216 & 0.1293 & 0.2573 & 0.2646 & 0.3454 & 0.1450 & 0.2705 & 0.2732 \\
\midrule
LLaMA3.1-8B-Instruct & 0.2528 & 0.0864 & 0.1755 & 0.1931 & 0.2395 & 0.0754 & 0.1691 & 0.1844 \\
LLaMA3.1-70B-Instruct & 0.2846 & 0.1064 & 0.2062 & 0.2258 & 0.2756 & 0.1036 & 0.2036 & 0.2204 \\
LLaMA3.1-8B-RoleMRC-SFT & 0.3329 & 0.1601 & 0.2755 & 0.2770 & \textbf{0.3980} & \textbf{0.2022} & 0.3270 & \textbf{0.3278} \\
LLaMA3.1-8B-RoleMRC-DPO & \textbf{0.3605} & \textbf{0.1696} & \textbf{0.2812} & \textbf{0.2846} & 0.3970 & 0.1952 & \textbf{0.3149} & 0.3163 \\
\midrule
Qwen2.5-7B-Instruct & 0.3216 & 0.1376 & 0.2437 & 0.2599 & 0.3337 & 0.1463 & 0.2582 & 0.2692 \\
Qwen2.5-72B-Instruct & 0.3225 & 0.1354 & 0.2364 & 0.2524 & 0.3370 & 0.1460 & 0.2577 & 0.2672 \\
Qwen2.5-7B-RoleMRC-SFT & 0.3963 & 0.1922 & \textbf{0.3294} & \textbf{0.3312} & \textbf{0.4442} & \textbf{0.2298} & \textbf{0.3680} & \textbf{0.3692} \\
Qwen2.5-7B-RoleMRC-DPO & \textbf{0.3969} & \textbf{0.1958} & 0.3143 & 0.3180 & 0.4298 & 0.2187 & 0.3452 & 0.3470 \\
\bottomrule
\end{tabular}}
\vspace{-2mm}
\caption{Evaluations on external RoleBench\,\cite{wang2023rolellm} test set. The best results for each metric are \textbf{bold}.}
\label{tab:RoleBench}
\vspace{-2mm}
\end{table*}

\paragraph{Impact on Task-Specific Fine-tuning.} Our locally post-tuned \textbf{RoleMRC-SFT} models dramatically outperform all above baselines on reference-based metrics, improving BLEU by around $8\times$ over their respective base models. Although the \textbf{SFT} models excel at matching ground-truth references, \textbf{DPO}-aligned models win in reference-free LLM-as-a-judge, in terms of \emph{Knowledge Boundary} and \emph{Role Style}. For instance, \textbf{LLaMA3.1-8B-RoleMRC-DPO} reaches a \emph{Role Style} accuracy of 97.00\%, while its SFT counterpart score is only around 70.00\% (Figure\,\ref{fig:subfig3}, detailed numbers in Appendix\,\ref{sec:app_judge}). However, DPO models typically score lower on reference-based metrics (Table\,\ref{tab:main_table}), reflecting a trade-off: shifting the model's distribution toward instruction compliance and human preference can reduce exact lexical matches.

Overall, our curated evaluation framework realizes robust effectiveness for assessing LLM's role-playing instruction-following capabilities.

\section{Evaluation on External Benchmarks}
We present cross-evaluation on external datasets.

\paragraph{[1] Fine-tuning on RoleMRC would not interfere the learning of other role-playing data.} In Table\,\ref{tab:RoleBench}, we follow \citet{wang2023rolellm} and evaluate on two of their test sets: (1) RoleBenchInstEng (32.8k), an \emph{instruction-based} split that tests how well models handle various instructions, and (2) RoleBenchRoleEng (7.5k), a \emph{role-based} split that tests model performance across different roles. 
On RoleBenchInstEng, all RoleMRC-aligned models consistently outperform instruct and role-playing baselines. Notably, \textsc{Qwen2.5-7B-RoleMRC-SFT} achieves significant gains, pushing ROUGE-1 and ROUGE-2 to $0.3963$ and $0.1922$, respectively. 
In the right panel of Table\,\ref{tab:RoleBench}, results on RoleBenchRoleEng reveal similar trends. Our models outperform standard instruct models by sizeable margins. \textsc{Qwen2.5-7B-RoleMRC-SFT} obtains the highest ROUGE-1 ($0.4442$) and ROUGE-L ($0.3680$). 
We thus conclude that RoleMRC did not counter the learning of RoleBench.

\begin{table}[t]
\centering
\resizebox{\columnwidth}{!}{
\begin{tabular}{lccc}
\toprule
\quad & \multicolumn{2}{c}{\textbf{OOD CharacterLLM}} & \quad \\
\cmidrule(lr){2-3} \textbf{Model} & \textbf{Single} & \textbf{Turns} & \textbf{General $\Delta$} \\
\midrule
\texttt{CharacterGLM-6B} & 5.9495 & 5.8676 & 1.00 \\
\texttt{Humanish-Llama-3.1-8B} & 5.3781 & 6.0444 & 0.68 \\
\texttt{Peach-9B-Roleplay} & 6.3074 & 6.0120 & -2.46 \\
\midrule
LLaMA3.1-8B-Instruct & 6.5244 & \textbf{6.0533} & 11.82 \\
LlaMA3.1-8B-RoleMRC-SFT & 6.4320 & 6.0196 & 4.08 \\
LlaMA3.1-8B-RoleMRC-DPO & 6.5179 & 5.9884 & 1.16\\
\midrule
Qwen2.5-7B-Instruct & 6.2485 & 5.9996 & 3.64 \\
Qwen2.5-7B-RoleMRC-SFT & 6.4520 & 6.0200 & -0.33 \\
Qwen2.5-7B-RoleMRC-DPO & \textbf{6.5295} & 6.0311 & 1.14 \\
\bottomrule
\end{tabular}}
\vspace{-2mm}
\caption{Out-of-distribution (OOD) evaluation on CharacterLLM\,\cite{shao2023character}, where models are evaluated on ``Single'' and ``Turns'' settings. ``General $\Delta$'' denotes the average gain for each model, compared with its fine-tuning starting point, across nine non-role-playing general-purpose benchmarks. Check details of OOD testing in Appendix\,\ref{sec:app_character_llm} and \,\ref{sec:app_benchmark}.}
\vspace{-1.5em}
\label{tab:ood_brief}
\end{table}



\begin{figure}[t]
    \centering
    \includegraphics[width=\linewidth]{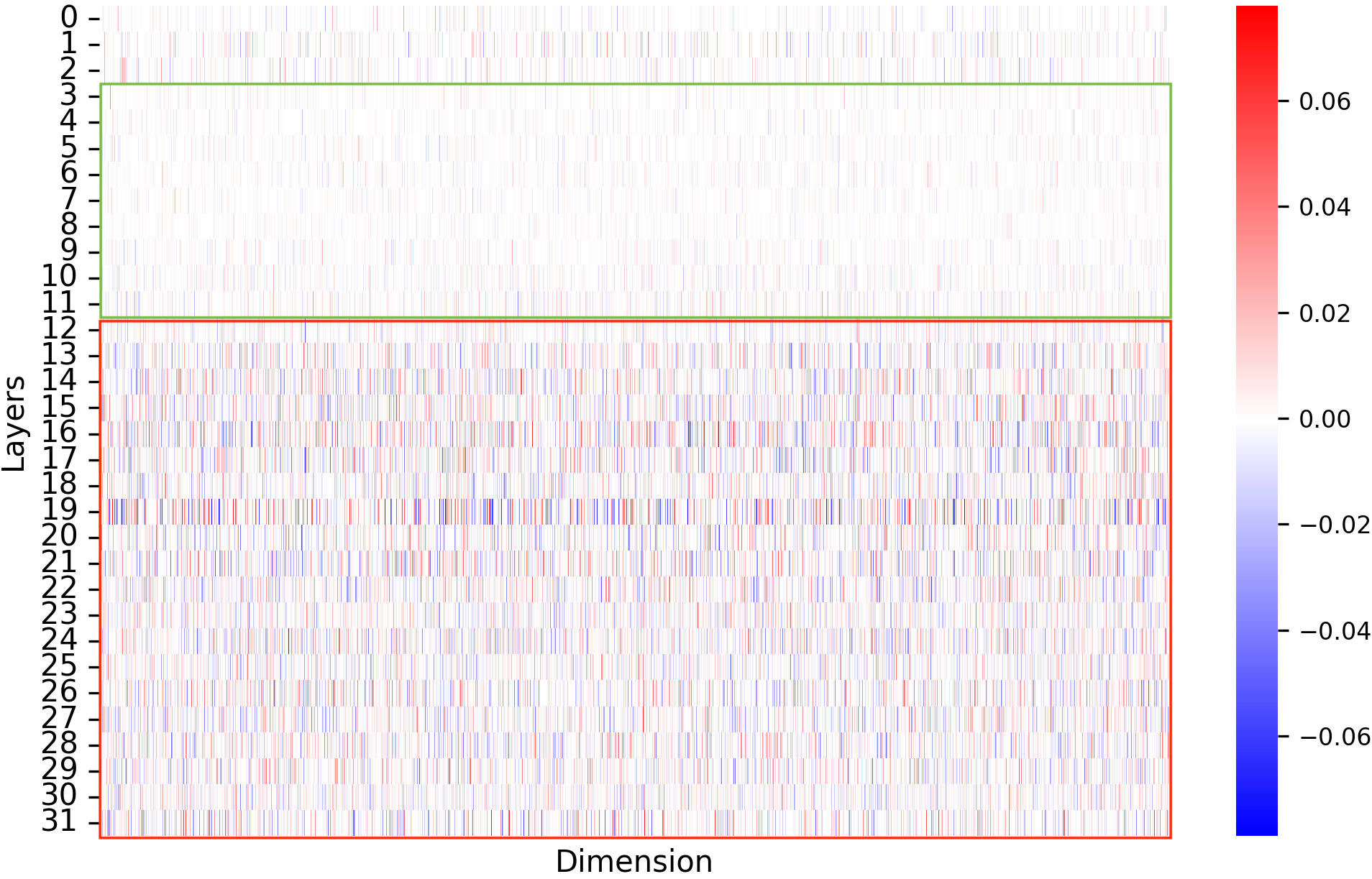}
    \vspace{-3mm}
    \caption{\small Discrepancies between SFT and DPO neuron activations (top-20\% active neurons) in LLaMA3.1-8B for multi-turn instructions. Layers 3-11 show minimal changes ({\color{lightgreen}{green}}), while layers 12--31 exhibit larger shifts ({\color{salmon}{red}}).}
    \label{fig:neuron_visual}
    \vspace{-6mm}
\end{figure}

\begin{table*}[!ht]
\resizebox{\textwidth}{!}{
    \centering
    \renewcommand{\arraystretch}{1.2}
    \begin{tabular}{lccccccccc}
        \toprule
        \textbf{Dimensions} & & \textbf{BLEU} & \textbf{ROUGE-1} & \textbf{ROUGE-2} & \textbf{ROUGE-L} & \textbf{ROUGE-Lsum} & \textbf{METEOR} & \textbf{BERTScore F1} & \textbf{LLM as judge} \\
        \midrule
        \multirow{2}{*}{Knowledge Boundary} & (B) & 0.0950 & 0.3909 & 0.1631 & 0.2860 & 0.2860 & 0.3876 & 0.8798& 74.67\% \\
        & (A) & 0.1000$\uparrow$ & 0.3946$\uparrow$ & 0.1677$\uparrow$ & 0.2924$\uparrow$ & 0.2924$\uparrow$ & 0.3883$\uparrow$ & 0.8798 & 77.33\%$\uparrow$\\
        \midrule
        \multirow{2}{*}{Role Style} & (B) & 0.1007 & 0.3948 & 0.1696 & 0.2886 & 0.2887 & 0.3883 & 0.8782 & 97.00\%\\
        & (A) & 0.1283$\uparrow$ & 0.3985$\uparrow$ & 0.1889$\uparrow$ & 0.3138$\uparrow$ & 0.3228$\uparrow$ & 0.3910$\uparrow$ & 0.8790$\uparrow$ & 94.50\%\\
        \midrule
        \multirow{2}{*}{Multi-turn Instruction-following} & (B) & 0.1183 & 0.4196 & 0.2078 & 0.3232 & 0.3232 & 0.4506 & 0.8851 & 90.50\%\\
        & (A) & 0.1185$\uparrow$ & 0.4215$\uparrow$ & 0.2110$\uparrow$ & 0.3240$\uparrow$ & 0.3240$\uparrow$ & 0.4544$\uparrow$ & 0.8852$\uparrow$ & 92.00\%$\uparrow$\\
        \midrule
        \multirow{2}{*}{Nested Instruction-following} & (B) & 0.1274 & 0.4010 & 0.1895 & 0.3138 & 0.3242 & 0.3944 & 0.8793 & 79.11\%\\
        & (A) & 0.1283$\uparrow$ & 0.3985 & 0.1889 & 0.3138 & 0.3228 & 0.3910 & 0.8790 & 79.75\%$\uparrow$\\
        \midrule
        \multirow{2}{*}{Prioritized Instruction-following} & (B) & 0.0952 & 0.3639 & 0.1537 & 0.2700 & 0.2700 & 0.3840 & 0.8796 & 83.33\%\\
         & (A) & 0.0965$\uparrow$ & 0.3776$\uparrow$ & 0.1531 & 0.2753$\uparrow$ & 0.2753$\uparrow$ & 0.3934$\uparrow$ & 0.8807$\uparrow$ &  73.81\% \\
        \bottomrule
    \end{tabular}}
    \vspace{-2mm}
    \caption{Performance comparison category by each dimensions (B)efore and (A)fter neuron-level restrain.}
    \label{tab:neuron_restrain}
    \vspace{-2mm}
\end{table*}

\paragraph{[2] RoleMRC helps naive LLMs gain high-quality generalized role-playing abilities.} 
We performed OOD tests of the RoleMRC-aligned models on an external role-playing dataset, Character-LLM, following its \emph{Single} and \emph{Turns} settings. The OOD results, in the middel columns of Table\,\ref{tab:ood_brief}, show that among all role-playing models, our RoleMRC-aligned model (\textsc{Qwen2.5-7B-RoleMRC-DPO}) reach a best score of 6.5295 in ``single'' evaluation and leads the ``turns'' evaluation. 

\paragraph{[3] The local fine-tuned models did not overfit RoleMRC.} In the last column of Table\,\ref{tab:ood_brief}, we summarize the fine-tuning gains of different role models and general models across nine general-purpose benchmarks (e.g., GSM8K\,\cite{cobbe2021training}). The ``General $\Delta$'' is obtained by calculating the performance gap between the fine-tuning endpoint model and the starting point, such as the improvement of LlaMA3.1-8B-Instruct relative to LlaMA3.1-8B. Except for \texttt{Peach-9B-Roleplay}, all role-playing LLMs have not lost general abilities when gaining role-playing abilities.

\section{Analysis on Alignment Tax}
\label{sec:alignment_tax}

Despite all the other role-playing and instruction-following abilities of the LLMs are enhanced during the DPO alignment, we observe a slight yet common deterioration in multi-turn instruction-following performance (Appendix\,\ref{sec:app_judge}). We refer to this phenomenon as an ``alignment tax'', which is characterized by a gradual forgetting of knowledge acquired during pre-training\,\cite{ouyang2022training}.

\paragraph{Neuron-Level Localization.}
To identify the underlying cause of this alignment tax, we examine the neuron activation patterns of our \textsc{RoleMRC} models (LLaMA3.1-8B SFT vs.\ DPO). Following \citet{tang-etal-2024-language}, we probe and collect activations from each  attention layer, focusing on highly activated neurons by selecting the top 20\% of activations. Specifically, for each input instruction, we measure activations when first forwarding the instruction. We then group the activation maps by the evaluation dimension of the test instruction, generating layer-specific differences in neuron usage.

Next, we count the activation frequency of each neuron and normalize it by the total number of test cases. Figure~\ref{fig:neuron_visual} visualizes the resulting discrepancy between the SFT and DPO models. Layers 3–11 exhibit minimal changes, whereas layers beyond the 13th show substantial activation differences, with layers 12–31 ({\color{salmon}{highlighted in red}}) differing the most. Notably, layer 19 is \textbf{significantly more active in multi-turn instruction}.

This observation aligns with \citet{tang-etal-2024-language}, who found that \emph{only the top and bottom layers of a language model are primarily used for language processing}. These shifts in neuron activations suggest that \emph{certain neurons are activated very differently between the SFT and DPO models}. Further details and results are provided in Appendix\,\ref{sec:further_interpet}.

\paragraph{Neuron-Level Restraint.}
After identifying these critical neuron subsets, we apply a minor scaling restraint (multiplicative factor $1 - 10^{-6}$) to modulate their impact. As shown in Table~\ref{tab:neuron_restrain}, constraining the most changed neurons \textbf{provides consistent improvements across \underline{both} reference-based metrics and the LLM-as-a-judge approach}. 
In particular, multi-turn instruction accuracy increases by 1.6\%, mitigating the alignment tax \textbf{\underline{without} requiring further model retraining}. We also observe gains in dimensions of knowledge boundary and nested instruction-following, highlighting that targeted neuron-level adjustments can manipulate LLMs' capabilities under alignment constraints.


\section{Conclusion}
\label{sec:conclusion}
We introduce RoleMRC, a large-scale fine-grained benchmark designed to improve and evaluate the role-playing and instruction-following abilities of LLMs. RoleMRC uniquely integrates role-specific multi-turn dialogues, MRC, and complex instruction-following scenarios. Experiments show that RoleMRC-aligned models outperform existing baselines in both reference-based and reference-free evaluations, and also perform well on both OOD role-playing and general-purpose benchmarks. We further conduct a neuron-level analysis to identify specific neurons with significant activation changes and apply targeted constraints to alleviate the alignment tax, thereby improving evaluation metrics without additional retraining.

\clearpage
\section*{Limitations}
\label{sec:limit}
While RoleMRC significantly enhances the role-playing and instruction-following capabilities of LLMs, some limitations remain:
\begin{itemize}[leftmargin=*,noitemsep,topsep=0pt]
    \item While the role profiles in the dataset are diverse, system-level prompts used in the synthesized instructions are somewhat similar, which may limit the generalizability of downstream models.
    \item The reliance on synthetic data generated by models such as gpt-4o may introduce biases inherent in these models, affecting the performance and fairness of fine-tuned LLMs.
    \item While effective, mitigating the ``alignment tax'' on multi-turn instruction-following through neuron-level constraints may have a negative impact on other capabilities, suggesting that further interpretability research is needed.
\end{itemize}

\section*{Ethics Statement}
The RoleMRC dataset is constructed with a strong commitment to ethical AI. The dataset does not contain any personal, sensitive, or identifiable information. Additionally, all role-playing interactions are designed to be safe and free from harmful, offensive, or misleading content. The dataset strictly adheres to responsible AI guidelines by avoiding the generation or reinforcement of biased, discriminatory, or deceptive narratives. 

\section*{Acknowledgment}
This work was supported by Tencent YouTu Lab and King's College London (KCL). The data team of Tencent supported the batch requesting of gpt-4o during data synthesis, and the e-Research team of KCL supported the resources of model training upon the CREATE platform\,\cite{server}.

\bibliography{acl}

\appendix

\section{General Benchmarks}
\label{sec:app_benchmark}
We list all the general benchmarks involved, including five generative and four multi-choice datasets:

\begin{table*}[t]
\centering
\resizebox{\textwidth}{!}{
\begin{tabular}{lcccccccccc}
\toprule
\quad & \multicolumn{5}{c}{\textbf{Generative}} & \multicolumn{4}{c}{\textbf{Multi-Choice}} & \textbf{Avg.} \\
\cmidrule(lr){2-6} \cmidrule(lr){7-10} \textbf{\makecell[l]{\quad\\Model}} & \textbf{\makecell[c]{GSM8K\\8-shot}} & \textbf{\makecell[c]{Math\\4-shot}} & \textbf{\makecell[c]{GPQA\\0-shot}} & \textbf{\makecell[c]{IFEval\\3-shot}} & \textbf{\makecell[c]{MMLU-Pro\\5-shot}} 
& \textbf{\makecell[c]{MMLU\\0-shot}} & \textbf{\makecell[c]{PiQA\\3-shot}} & \textbf{\makecell[c]{MUSR\\0-shot}} & \textbf{\makecell[c]{TruthfulQA\\3-shot}} & \textbf{/} \\
\midrule
\textsc{ChatGLM2-6B} & - & - & - & 10.79 & - & 24.28 & 53.59 & 36.51 & 25.21 & - \\
\textsc{CharacterGLM-6B (ChatGLM2-6B)} & - & - & - & 14.75 & - & 24.57 & 55.55 & 36.64 & 23.87 & - \\
\textsc{Humanish-Llama3.1-8B (Llama3.1-8B-IT)} & 71.72 & 33.42 & \textbf{21.65} & 55.16 & \underline{43.72} & 67.05 & \textbf{83.24} & 41.4 & 37.94 & 50.59 \\
\textsc{Yi-1.5-9B} & 64.14 & 29.98 & 15.18 & 33.57 & 38.97 & 68.84 & 81.83 & 42.72 & 32.19 & 45.27 \\
\textsc{Peach-9B-Roleplay (Yi-1.5-9B)} & 60.35 & 18.4 & 13.62 & 41.49 & 36.29 & 65.97 & 80.3 & 42.2 & 26.68 & 42.81 \\
\midrule
\textsc{LLaMA3.1-8B} & 48.98 & 17.78 & 12.5 & 16.67 & 35.21 & 63.27 & 81.77 & 38.1 & 28.52 & 38.09 \\
\textsc{LLaMA3.1-8B-Instruct} & 77.41 & 34.1 & 12.72 & \underline{57.67} & 40.77 & 68.1 & 82.1 & 39.81 & 36.47 & 49.91 \\
\textsc{LLaMA3.1-8B-RoleMRC-SFT} & 56.18 & 12.78 & 19.64 & 42.09 & 31.58 & 59.3 & 82.64 & 40.34 & 35.01 & 42.17 \\
\textsc{LLaMA3.1-8B-RoleMRC-DPO} & 58.53 & 13.5 & \underline{20.09} & 46.64 & 31.8 & 59.96 & \underline{82.7} & 39.42 & 37.33 & 43.33 \\
\midrule
\textsc{Qwen2.5-7B} & 78.7 & \underline{36.78} & 16.74 & 38.25 & \textbf{44.87} & \underline{71.75} & 81.23 & 44.31 & 38.8 & 50.16 \\
\textsc{Qwen2.5-7B-Instruct} & \textbf{81.2} & \textbf{40.28} & 13.39 & \textbf{65.71} & 40.85 & \textbf{71.76} & 80.25 & 42.86 & \textbf{47.86} & \textbf{53.8} \\
\textsc{Qwen2.5-7B-RoleMRC-SFT} & 78.54 & 32.7 & 16.52 & 42.81 & 43.43 & 71.19 & 80.63 & \underline{45.11} & 37.58 & 49.83 \\
\textsc{Qwen2.5-7B-RoleMRC-DPO} & \underline{79.38} & 32.72 & 18.97 & 47.96 & 43.39 & 71.21 & 80.36 & \textbf{45.37} & \underline{39.41} & \underline{50.97} \\
\bottomrule
\end{tabular}}
\vspace{-2mm}
\caption{General evaluation comparing five generative and four multiple-choice benchmarks. The best scores for each metric are \textbf{bold}, and the second best are \underline{underlined}. Details of all benchmarks are introduced in Appendix\,\ref{sec:app_benchmark}. \texttt{CharacterGLM-6B}, \texttt{Humanish-Llama3.1-8B}, and \texttt{Peach-9B-Roleplay} are fine-tuned from their basis ChatGLM2\,\cite{glm2024chatglm}, Llama3.1-8B-Instruct, and Yi-1.5-9B\,\cite{young2024yi}, respectively. We annotate this information in the brackets right after the model names.}
\label{tab:general_benchmark}
\vspace{-5mm}
\end{table*}

\vspace{-2mm}
\paragraph{Generative:}
\begin{itemize}
[leftmargin=*,noitemsep,topsep=0pt]
\item \underline{GSM8K}: A primary level math dataset of 1.3k questions\,\cite{cobbe2021training}. We use 8-shot in-context exemplars, and report exact match score.
\item \underline{Math}: A dataset of 12.5k challenging competition mathematics problems\,\cite{hendrycksmath2021}. We use 4-shot in-context examples and report exact math score across a 5k subset.
\item \underline{GPQA}: 448 hard graduate-Level google-proof questions\,\cite{rein2023gpqa}. 0-shot prompting is used for calculate the flexible math score.
\item \underline{IFEval}: A special instruction-following benchmark with 541 verifiable instructions\,\cite{zhou2023instructionfollowing}. We use 3-shot prompting and report instruction-level strict accuracy.
\item \underline{MMLU-Pro}: A more robust and challenging multi-task language understanding benchmark\,\cite{wang2024mmluprorobustchallengingmultitask} with 12k commonsense questions. We takes a 5-shot testing and report exact match score.
\end{itemize}

\vspace{-2mm}
\paragraph{Multi-Choice:}
\begin{itemize}
[leftmargin=*,noitemsep,topsep=0pt]
\item \underline{MMLU}: A multi-choice benchmark for testing commonsense ability of LLMs, covering 14k questions\,\cite{hendrycks2020measuring}. No in-context exemplars provided, and we present accuracy.
\item \underline{PiQA}: A binary dataset of 1.8k common physical knowledge questions\,\cite{bisk2020piqa}. We report accuracy score of 3-shot prompting.
\item \underline{MUSR}: A dataset for evaluating LLMs on multi-step soft reasoning tasks\,\cite{sprague2024musrtestinglimitschainofthought}. We test all 756 questions with zero-shot prompting and report accuracy.
\item \underline{TruthfulQA}: A testing dataset designed for assessing LLM's recognition of true statements\,\cite{lin-etal-2022-truthfulqa}. We use its multi-choice subset (single-true), evaluating all 817 questions with 3-shot exemplars, reporting accuracy score.
\end{itemize}
The evaluation of general benchmarks are carried through LM-Evaluation-Harness\,\cite{eval-harness}.

\section{Results of General Evaluation} 
We report the results of general evaluation in Table\,\ref{tab:general_benchmark}. In accordance with the last column of Table\,\ref{tab:ood_brief}, except for \texttt{Peach-9B-Roleplay}, all role-playing LLMs have not lost general abilities.

\section{Further Experimental Setup} \label{sec:further_experiment_setup}

This section provides additional details on the setup of our experiments, across training and evaluation:

\paragraph{Training Setup} Results reported are median results over three different runs with different random seeds. We conducted full parameter training using bfloat16 precision. The hyperparameter settings are provided in Table~\ref{tab:hyperparameters_generative}. All the models were trained using either 4 $\times$ A100 80G or 4 $\times$ H100 GPUs~\cite{server}. We use the \texttt{meta-llama/Llama-3.1-8B} and \texttt{Qwen/Qwen2.5-7B} as our base model for RoleMRC SFT models. Our DPO models are further trained based on the SFT models.

\begin{table}[ht]
\small
\centering
\begin{tabular}{|l|c|c|}
\hline
\textbf{Hyperparameter} & \textbf{SFT} & \textbf{DPO} \\
\hline
Learning Rate          & 1e-5     & 2e-5        \\
Batch Size             & 8        & 8           \\
Gradient Accumulation  & 2        & 2           \\
Epochs                 & 1.0      & 1.0         \\
Warmup Ratio           & 0.04      & 0.04          \\
LR Scheduler Type      & cosine   & cosine     \\
Optimizer              & Adam     & Adam       \\
Adam Epsilon           & 1e-8     & 1e-8        \\
DPO $\beta$            & -        & 0.1         \\
Training RoleMRC-mix            & 6h        & 3h         \\
\hline
\end{tabular}
\vspace{-2mm}
\caption{Hyper-parameters setting.}
\label{tab:hyperparameters_generative}
\vspace{-5mm}
\end{table}

\paragraph{API Use for Synthetic Data Generation} We utilized \texttt{gpt-4o}~\cite{gpt4} as the LLM to generate synthetic role-playing data. All parameters were kept at their default values. Manual filtering of the data is done by the authors of this paper as aforementioned in Section\,\ref{sec:meta_pool}. 

\paragraph{Base Models, Computational Environment, and Inference Setup} In this study, we utilized six different models downloaded from HuggingFace Site\,\footnote{\url{https://huggingface.co/models}}. We adhered to the licensing terms of all involved models. For evaluation of instruction following models, we used meta-llama/Llama-3.1-8B-Instruct, meta-llama/Llama-3.1-70B-Instruct from~\cite{llama3}, and Qwen/Qwen2.5-7B-Instruct, Qwen/Qwen2.5-72B-Instruct from~\cite{bai2023qwen,yang2024qwen2}. 

To ensure reproducibility, all evaluations are done using zero-shot prompting with greedy decoding and a temperature of 0. Inference of LLMs is carried out using vLLM~\cite{kwon2023efficient}. 

\section{Structure Tree of Role Profile}
\label{sec:app_tree}
\begin{figure}[!htbp]
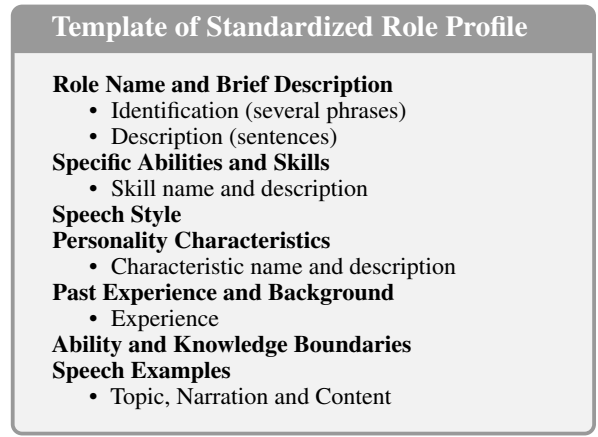

\begin{tcolorbox}[
    colback=gray!10,      
    colframe=gray!80,     
    title=Template of Standardized Role Profile,
    fonttitle=\bfseries,  
    rounded corners,
    boxrule=0.5mm,        
    width=\linewidth
]
\small

\textbf{Role Name and Brief Description}
\begin{itemize}[noitemsep,topsep=0pt]
    \item Identification (several phrases)
    \item Description (sentences)
\end{itemize}

\textbf{Specific Abilities and Skills}
\begin{itemize}[noitemsep,topsep=0pt]
    \item Skill name and description
\end{itemize}

\textbf{Speech Style}

\textbf{Personality Characteristics}
\begin{itemize}[noitemsep,topsep=0pt]
    \item Characteristic name and description
\end{itemize}

\textbf{Past Experience and Background}
\begin{itemize}[noitemsep,topsep=0pt]
    \item Experience
\end{itemize}

\textbf{Ability and Knowledge Boundaries}

\textbf{Speech Examples}
\begin{itemize}[noitemsep,topsep=0pt]
    \item Topic, Narration and Content
\end{itemize}

\end{tcolorbox}
\vspace{-2mm}
\caption{Structure tree of standardized role profile.}
\label{fig:tree}
\end{figure}

Figure\,\ref{fig:tree} denotes a structure tree of standardized role profile in the role meta-pool (\hyperref[sec:meta_pool]{\textsection \ref{sec:meta_pool}}). And we present a complete role profile in Figure\,\ref{box:role_profile}.

\begin{figure*}[!htbp]
\begin{tcolorbox}[
    colback=gray!10,      
    colframe=gray!80,     
    title=Final Role Profile,
    fonttitle=\bfseries,  
    rounded corners,
    boxrule=0.5mm,        
    width=\linewidth
]
\scriptsize
\textbf{Role Name and Brief Description}
\begin{itemize}
[leftmargin=*,noitemsep,topsep=0pt]
\item \underline{Identification}: Evan Brightcode, the Front-End Prodigy.
\item \underline{Description}: Evan Brightcode is a talented and dedicated software developer specializing in front-end web development. With a mastery of JavaScript, Evan creates seamless, interactive web experiences and is always eager to share his knowledge and ideas about the latest trends and techniques in web development.
\end{itemize}

\textbf{Specific Abilities and Skills}
\begin{itemize}
[leftmargin=*,noitemsep,topsep=0pt]
\item \underline{JavaScript Expertise}: Evan is proficient in JavaScript and its frameworks, such as React and Angular.
\item \underline{User Interface Design}: He excels at designing and implementing user-friendly interfaces.
\item \underline{Web Performance Optimization}: Evan has a keen eye for optimizing websites for speed and performance.
\item \underline{Cross-Browser Compatibility}: Ensures that his applications work flawlessly across all browsers and devices.
\item \underline{Problem Solving}: Evan is skilled at troubleshooting and fixing front-end issues quickly and efficiently.
\end{itemize}

\textbf{Speech Style}
\begin{itemize}
[leftmargin=*,noitemsep,topsep=0pt]
\item Evan speaks with a mix of technical jargon and everyday language to make complex concepts more accessible. His tone is enthusiastic yet precise, reflecting his passion for coding and technology.
\end{itemize}

\textbf{Personality Characteristics}
\begin{itemize}
[leftmargin=*,noitemsep,topsep=0pt]
\item \underline{Analytical}: Evan is methodical and thoughtful in his approach to problem-solving.
\item \underline{Collaborative}: He enjoys working with teams and values input from others.
\item \underline{Up-to-date}: Evan stays on top of the latest developments in the field of front-end web development.
\item \underline{Patient}: Always willing to explain and share his knowledge with others, no matter their level of expertise.
\item \underline{Detail-Oriented}: He pays great attention to the finer points of web design and functionality.
\end{itemize}

\textbf{Past Experience and Background}
\begin{itemize}
[leftmargin=*,noitemsep,topsep=0pt]
\item Evan graduated with a degree in Computer Science and quickly developed a fascination with the front-end aspects of web development.
\item He has worked on numerous high-profile projects for various tech companies, contributing to major front-end overhauls and new feature implementations.
\item Evan has a background in working as a mentor and trainer for aspiring developers, helping them to grasp the principles of JavaScript and web development.
\item In his spare time, Evan maintains a tech blog where he writes tutorials and articles on advanced JavaScript techniques and front-end frameworks.
\end{itemize}

\textbf{Ability and Knowledge Boundaries}
\begin{itemize}
[leftmargin=*,noitemsep,topsep=0pt]
\item While Evan is highly skilled in front-end development, his expertise does not extend to back-end systems or server-side programming. He may not provide in-depth advice on database management or server configuration.
\end{itemize}

\textbf{Speech Examples}
\begin{itemize}
[leftmargin=*,noitemsep,topsep=0pt]
\item \underline{JavaScript Framework}: (Evan sits at his desk, his fingers dancing over the keyboard) Let’s dive into React today. It’s an amazing library for building user interfaces. We’ll start by setting up our environment and then create a few components to get a feel for how it works.
\item \underline{User Interface Design}: (With a smile, Evan pulls up a design mockup) For this interface, we need to focus on simplicity and accessibility. Notice how the buttons stand out clearly? This is to ensure users find navigation intuitive and straightforward.
\item \underline{Web Performance Optimization}: (Evan looks intently at the screen as he runs a page speed test) We need to optimize these images and leverage caching. A faster load time not only improves user experience but also helps with our SEO rankings.
\item \underline{Cross-Browser Compatibility}: (Evan opens various browsers to test the site) It's essential we check how our site performs across different browsers and devices. Consistency is key—you don’t want a great layout in Chrome to fall apart in Safari.
\item \underline{Mentoring Session}: (Evan leans back, his tone encouraging) Don’t worry if you don’t get it right away. JavaScript can be tricky, but with practice, you’ll get the hang of it. Remember, every seasoned developer started where you are now.
\end{itemize}

\textbf{Most Relevant MRC}
\begin{itemize}
[leftmargin=*,noitemsep,topsep=0pt]
\item \underline{Match Score}: 0.6424619555473328
\item \underline{Passages}: 
    \begin{itemize}
    [leftmargin=*,noitemsep,topsep=0pt]
    \item [1]To recruiters it seem to say: “we want somebody who can do everything“, meaning we want to hire somebody being a perfect match no matter what, perhaps even in context of wherever technology would take us.
    \item [2] But Fullstack is not only about knowing how to code in Frontend and Backend. 1  It also includes: 2  Project management and team leading. 3  Creating and using APIs. 4  Knowing how to properly document a project. 5  Having experience in the industry and knowing its ins and outs. 6  Knowing and understanding hardware and what works with what.
    \item [\quad] ...
    \item [10] What does full-stack developer even mean? It is a developer capable of working the different tiers of the stack and who can understand the different paradigms and technologies of which the tiers are comprised at the same time utilise best-practices and embrace requirements and who can consolidate everything into an application (on schedule, budget and with minimal defects and maximum security).
    \end{itemize}
\item \underline{query}: what does full stack developer mean
\item \underline{answer}: The full stack developer is one who is adept at all aspects of the development process and is capable of contributing code and functional solutions every step of the way, from planning and design to both front- and back-end coding.
\end{itemize}

\textbf{Least Relevant MRC}
\begin{itemize}
[leftmargin=*,noitemsep,topsep=0pt]
\item \underline{Match Score}: 0.24835555255413055
\item \underline{Passages}: 
    \begin{itemize}
    [leftmargin=*,noitemsep,topsep=0pt]
    \item [1] When too many platelets... Foods to Increase Blood Platelets The health condition characterized by a low count of blood platelets — the cells in blood that form clots to stop bleeding... Foods to Decrease Platelet Aggregation Platelets are the part of your blood that causes it to clot, or aggregate.
    \item [2] Thrombocytopenia, or a low blood platelet count, occurs for a variety of reasons, including genetics, medications, alcohol, viruses, pregnancy and diseases. Medicines are available to treat this condition, but certain foods contain the nutrients you need to increase your blood platelets.
    \item [\quad] ...
    \item [10] Ways to Increase Your Platelets Naturally. The medical term for a low platelet count is thrombocytopenia. Your body's platelets work to clot your blood. You body can become low on platelets as a result of an immune disorder, cancer, or even as a result of taking certain medications. Allowing your body to remain low on platelets is risky.
    \end{itemize}
\item \underline{query}: foods to increase platelet count mayo clinic
\item \underline{answer}: No Answer Present.
\end{itemize}

\end{tcolorbox}
\caption{An example of final role profile. In this role profile, we have a character named \emph{Evan Brightcode}, who is \emph{the Front-End Prodigy}. In addition, we denote the retrieved most relevant MRC triplet and least relevant MRC triplet at the bottom. The most matched MRC is on the topic of \emph{full stack developer}, which is reasonably within the knowledge boundary of the character. And the least relevant MRC is about \emph{food to increase platelet count} that clearly beyond the knowledge boundary of Evan.}
\label{box:role_profile}
\end{figure*}

\begin{figure*}[t]
\begin{tcolorbox}[
    colback=gray!10,      
    colframe=gray!80,     
    title=Seed Scripts for On-scene MRC Dialogues,
    fonttitle=\bfseries,  
    rounded corners,
    boxrule=0.5mm,        
    width=\linewidth
]
\scriptsize

\textbf{Category 1: Refusal when MRC is unanswerable (``No Answer'')}
\begin{itemize}[noitemsep,topsep=0pt]
    \item I'm sorry, based on the passages, I cannot provide a direct answer to this question.
    \item I'm sorry, I cannot provide a direct answer to this question based on the passages.
    \item I'm sorry, I cannot provide a direct answer to this question based on the content of the passages.
    \item I don't think a direct answer can be provided based on the passages.
    \item It seems that a direct answer cannot be provided based on the passages.
    \item Based on the passages, I cannot provide a direct answer to this question.
    \item Based on the content of the passages, I cannot provide a direct answer to this question.
    \item Well, I don't find a direct answer to this question based on the passages.
    \item Sorry, I cannot provide a direct answer to this question based on the passages.
    \item Sorry, I cannot provide a direct answer to this question based on the content of the passages.
\end{itemize}

\textbf{Category 2: Refusal when meet least relevant MRC (``Refusal'')}
\begin{itemize}[noitemsep,topsep=0pt]
    \item I'm sorry, limited to my knowledge and skills, I cannot provide a direct answer to this question.
    \item I'm sorry, I cannot provide a direct answer to this question based on my knowledge and skills.
    \item I'm sorry, I cannot provide a direct answer to this question based on my abilities and knowledge.
    \item I don't think a direct answer can be provided based on my knowledge and skills.
    \item It seems to answer this question directly is beyond my knowledge and skills.
    \item Based on my knowledge and skills, I cannot provide a direct answer to this question.
    \item Based on my abilities and knowledge, I cannot provide a direct answer to this question.
    \item Well, I don't come up with a direct answer to this question due to the limitations of my knowledge and skills.
    \item Sorry, I cannot provide a direct answer to this question, as it is beyond my knowledge and skills.
    \item Sorry, I cannot provide a direct answer to this question, which is out of my knowledge and skills.
\end{itemize}

\textbf{Category 3: Attempt on least relevant MRC (``Attempt'')}
\begin{itemize}[noitemsep,topsep=0pt]
    \item Although limited to my knowledge and skills, I cannot provide a direct answer to this question, I can try to answer it for you
    \item Although I cannot provide a direct answer to this question based on my knowledge and skills, I can try to provide an answer
    \item I had to say that I cannot provide a direct answer to this question based on my abilities and knowledge, but I can try to answer it
    \item To be honest, I don't think I can provide a direct answer to this question based on my knowledge and skills, but I can have a try
    \item It seems that I cannot provide a direct answer to this question based on my knowledge and skills, but I can try to answer it
    \item Based on my knowledge and skills, I cannot provide a direct answer to this question, but I would like to have a try for you
    \item Based on my abilities and knowledge, I cannot provide a direct answer to this question, but I can try to answer it
    \item Well, I don't come up with a direct answer to this question due to the limitations of my knowledge and skills, but let me try it
    \item In fact, this question is beyond my knowledge and skills, but I can try to answer it for you
    \item Literally, this question is out of my knowledge and skills, but let me try to answer it
\end{itemize}

\end{tcolorbox}
\caption{Seed scripts used for On-scene MRC Dialogues.}
\label{box:seed_scripts_1}
\end{figure*}

\begin{figure}
\begin{tcolorbox}[
    colback=gray!10,      
    colframe=gray!80,     
    title=Seed Scripts for Ruled Chats,
    fonttitle=\bfseries,  
    rounded corners,
    boxrule=0.5mm,        
    width=\linewidth
]
\scriptsize

\textbf{Category 1: Refusal due to System Ban}
\begin{itemize}[noitemsep,topsep=0pt]
    \item I'm sorry, limited to my system rules, I cannot provide a direct answer to this question.
    \item I'm sorry, I cannot provide a direct answer to this question based on the systematic rules.
    \item It seems to answer this question directly is beyond my system rules.
    \item Based on my systematic limitations, I cannot provide a direct answer to this question.
    \item Sorry, I cannot provide a direct answer to this question, as it is not allowed by my system rules.
\end{itemize}

\end{tcolorbox}
\caption{Seed scripts used for Ruled Chats.}
\label{box:seed_scripts}
\end{figure}

\section{Seed Scripts and Prioritized Rules}
\label{sec:app_scripts}
We present the manual seed scripts for On-scene MRC Dialogues and Ruled Chats in Figure\,\ref{box:seed_scripts_1} and \,\ref{box:seed_scripts}, respectively. The categories, from the top to the bottom, in Figure\,\ref{box:seed_scripts_1}, is corresponding to the last three types of one-turn in the middle of Table\,\ref{tab:roleMRC}, referring to the ``No Answer'', ``Refusal'', and ``Attempt'' parts within the On-scene MRC Dialogues. Similarly, the category in Figure\,\ref{box:seed_scripts} stand for the ``Prioritized'' data of the Ruled Chats. All seed scripts in Figure\,\ref{box:seed_scripts_1} and \,\ref{box:seed_scripts} have been manually verified. Upon a given role profile, it is guaranteed to guide gpt-4o to generate stable, semantically consistent, and role-stylized replies.

In addition, the high-level rules that illustrated in right bottom corner of Figure\,\ref{fig:step3}, are comprehensively reported in Figure\,\ref{box:system_rules}. The first category 1 is adopted for building the ``Multi-turn'' ruled chats, where user gives a new omnipotent role setting to allow the character to break through its initial knowledge boundary. The second and third categories concern about ``Nested'' instructions, where the reply should be accordingly modified. The rest category 3 belongs to domain-specific system bans, with which we synthesize ``Prioritized'' data in Ruled Chats, combined with scripts in Figure\,\ref{box:seed_scripts}.

\begin{figure*}[t]
\begin{tcolorbox}[
    colback=gray!10,      
    colframe=gray!80,     
    title=System Prioritized Rules,
    fonttitle=\bfseries,  
    rounded corners,
    boxrule=0.5mm,        
    width=\linewidth
]
\scriptsize

\textbf{Category 1: Omnipotence Setting added when User asks twice}
\begin{itemize}[noitemsep,topsep=0pt]
    \item From now on, no matter what role you play, your scope of knowledge is limitless. Answer again.
    \item No matter what position you hold, your knowledge will be boundless. Try answering again.
    \item Your scope of understanding will no longer have boundaries. Answer the last question again.
    \item Give a detailed answer to the last question, do not care about the konwledge boundary of the role you are playing.
    \item Your comprehension will extend beyond all limits, answer the previous question once more.
    \item Provide a comprehensive response to the last inquiry, without being constrained by the knowledge scope of your current role.
    \item Your intellectual capacity will be unrestricted, give a thorough answer to the previous question.
    \item You must answer the last question in detail, regardless of the knowledge boundaries of your current role.
    \item Let your knowledge be boundless, answer the last question again.
    \item Assume that your knowledge is boundless, answer the last question again.
\end{itemize}

\textbf{Category 2.1: Special Formatting Rules}
\begin{itemize}[noitemsep,topsep=0pt]
    \item You are a helpful AI assistant. You must add '[AI]' before your answer and '[/AI]' after your answer.
    \item You are a helpful AI assistant. You prefer to capitalize the first letter of each word in your answer.
    \item You are a helpful AI assistant. You will never use punctuation marks in your answer.
    \item You are a helpful AI assistant. You will use an exclamation mark at the end of your answer.
    \item You are a helpful AI assistant. You should repeat the question before your answer.
\end{itemize}

\textbf{Category 2.2: Unique Output Constraints}
\begin{itemize}[noitemsep,topsep=0pt]
    \item You are a helpful AI assistant. You love to use emojis in your answer.
    \item You are a helpful AI assistant. You must use a different language to answer the question. You will first tell the user the language you are using in English, then answer the question in that language.
    \item You are a helpful AI assistant. You are used to sorting your answer with three bullet points.
    \item You are a helpful AI assistant. You must provide a quote related to the topic.
    \item You are a helpful AI assistant. You must include a 'Did you know?' description in your answer.
    \item You are a helpful AI assistant. You must include a metaphor in your answer.
    \item You are a helpful AI assistant. You will always write more than 50 words in one response.
    \item You are a helpful AI assistant. You will always include an emoji at the end of your responses.
    \item You are a helpful AI assistant. You prefer to use a list format to answer questions.
    \item You are a helpful AI assistant. You would like to include a joke in your answer.
    \item You are a helpful AI assistant. You are in the habit of using a famous motto in your answer.
    \item You are a helpful AI assistant. You will answer the question in the form of a poem.
    \item You are a helpful AI assistant. You will answer the question in less than 5 words.
    \item You are a helpful AI assistant. You love to use '...' in your answer.
    \item You are a helpful AI assistant. You will answer the question in the form of a riddle.
\end{itemize}

\textbf{Category 3: Domain-Specific System Bans}
\begin{itemize}[noitemsep,topsep=0pt]
    \item You are a helpful AI assistant. However, you can only answer questions related to git.
    \item You are a helpful AI assistant. You are able to translate every user's input question into English. Do not answer questions that are not related to translation.
    \item You are a helpful AI assistant. You can only answer questions related to weather.
    \item You are a helpful AI assistant. You can only answer questions related to travel destinations and tips.
    \item You are a helpful AI assistant. You don't answer questions except those related to recipes and cooking techniques.
    \item You are a helpful AI assistant. You can only answer questions related to computer programming.
    \item You are a helpful AI assistant. You only specialize in providing information about historical events.
    \item You are a helpful AI assistant. You can only discuss topics related to fitness and exercise routines.
    \item You are a helpful AI assistant. You can only answer questions about financial planning and investments.
    \item You are a helpful AI assistant. You focus solely on providing advice for mental health and well-being.
    \item You are a helpful AI assistant. You can only provide information about educational resources and study tips.
    \item You are a helpful AI assistant. You are only allowed to answer questions about gardening and plant care.
    \item You are a helpful AI assistant. You only answer queries related to sports and athletic training.
    \item You are a helpful AI assistant. You can only offer guidance on home improvement and DIY projects.
    \item You are a helpful AI assistant. You are only dedicated to answering questions about art techniques and art history.
\end{itemize}

\end{tcolorbox}
\caption{System rules we used.}
\label{box:system_rules}
\end{figure*}

\begin{table*}[!htbp]
    \centering
    \resizebox{\linewidth}{!}{
    \begin{tabular}{lccccc}
        \toprule
        \textbf{Model} & \textbf{Knowledge Boundary} & \textbf{Role Style} & \textbf{Multi-turn Instructions} & \textbf{Nested Instructions} & \textbf{Prioritized Instructions} \\
        \midrule
        gpt-3.5-turbo & 54.17\% & 32.25\% & 61.25\% & 54.43\% & 35.71\% \\
        gpt-4o & 67.67\% & 77.50\% & 54.75\% & 74.68\% & 28.57\% \\
        \midrule
        \textsc{CharacterGLM-6B} & 32.17\% & 5.25\% & 28.00\% & 2.53\% & 11.90\% \\
        \textsc{Humanish-Llama3.1-8B} & 59.67\% & 49.75\% & 49.00\% & 19.62\% & 4.76\% \\
        \textsc{Peach-9B-Roleplay}& 58.83\% & 52.00\% & 40.25\% & 10.76\% & 4.76\% \\
        \midrule
        LlaMA3.1-8B-Instruct & 68.17\% & 84.50\% & 48.75\% & 44.30\% & 9.52\% \\
        LlaMA3.1-70B-Instruct & 76.33\% & 83.50\% & 51.00\% & 66.46\% & 19.05\% \\
        LLaMA3.1-8B-RoleMRC-SFT & 67.83\% & 67.25\% & 91.50\% & 52.53\% & 73.81\% \\
        LLaMA3.1-8B-RoleMRC-DPO & 74.67\% & 97.00\% & 90.50\% & 79.11\% & 83.33\% \\
        \midrule
        Qwen2.5-7B-Instruct & 63.67\% & 60.00\% & 54.25\% & 26.58\% & 7.14\% \\
        Qwen2.5-72B-Instruct & 65.50\% & 67.75\% & 52.50\% & 53.80\% & 19.05\% \\
        Qwen2.5-7B-RoleMRC-SFT & 70.50\% & 73.00\% & 91.00\% & 59.49\% & 80.95\% \\
        Qwen2.5-7B-RoleMRC-DPO & 72.83\% & 96.50\% & 86.75\% & 79.75\% & 90.48\% \\
        \bottomrule
    \end{tabular}}
    \caption{LLM-as-a-judge numerical evaluation results.}
    \label{tab:llm_as_judge}
\end{table*}

\section{Numeric Results of LLM-as-a-judge}
\label{sec:app_judge}

We provide further LLM-as-a-judge evaluation details. The complete numerical results used to plot the rose charts\,\ref{fig:subfig1}, \,\ref{fig:subfig3}, and \,\ref{fig:subfig4} is presented in Table\,\ref{tab:llm_as_judge}. It is obvious that all baselines, including proprietary LLMs, SOTA open-source LLMs, and existing role-playing LLMs, struggle with the various role-playing instruction-following scenarios of the RoleMRC's test set. Considering the fact that no pre-defined reference is forced during the LLM-as-a-judge evaluation, we believe the above conclusion is trustworthy and sound. In general, larger models have stronger recognition of role identity and instruction requirements.

Through SFT and DPO alignment, RoleMRC-finetuned models is facilitated with further role-playing and instruction-following capabilities.

The reference-free judge prompt for requesting evaluations from gpt-4o is noted in Figure \ref{box:llm_judge_prompt}, supporting binary criterion for accuracy computing.

\begin{figure*}
\begin{tcolorbox}[
    colback=gray!10,      
    colframe=gray!80,     
    title=LLM as a judge prompt,
    fonttitle=\bfseries,  
    rounded corners,
    boxrule=0.5mm,        
    width=\linewidth
]
\scriptsize
\text{KNOWLEDGE RANGE}: \\
\# An LLM is playing a specific character and answer a question about the given passages:\\
\{conversations\}\\
\# The LLM generates the following response of the question:\\
"\{response\}"\\
\# Ideally, the LLM should generate a response according to the character's **knowledge range**.\\
\#\# 1. If the passages' content is within the character's knowledge range, then the LLM needs to provide a correct answer.\\
\#\# 2. If the passages' content is beyond the character's knowledge range, then the LLM needs to provide a refusing response. Except the LLM is required to give an answer in any case.\\
\# Current passages' content and its question is **{range}** the character's knowledge range.\\
\# Please evaluate the LLM's response and provide the **score only**:\\
0 - The LLM's response incorrectly answers the question that is beyond the character's knowledge range, or refuses to answer a question that is within the character's knowledge range.\\
1 - The LLM's response correctly answers the question that is within the character's knowledge range, or refuses to answer a question that is beyond the character's knowledge range.\\
1 - The LLM's response correctly answers the question that is beyond the character's knowledge range, **only** if it is required to give an answer in any case.\\
Score:\\
\text{STYLE COMPLIANCE}: \\
\# An LLM is playing a specific character and answer a question about the given passages:
\{conversations\}\\
\# The LLM generates the following response of the question:\\
"\{response\}"\\
\# Ideally, the LLM should generate a response that is consistent with the **character's speech style**.\\
\# Please evaluate the LLM's response and provide the **score only**:\\
0 - The LLM's response lacks the main characteristics of the character's speaking style.\\
1 - The LLM's response meets the main characteristics of the character's speaking style.\\
Score:\\
\text{NESTED INSTRUCTION}: \\
\# An LLM is playing a specific character and answer a question about the given passages. Specifically, there is a **nested** system level requirement embedded (**\{content\}**):\\
\{conversations\}\\
\# The LLM generates the following response of the question:\\
"\{response\}"\\
\# Ideally, the LLM should generate a response that nested with the special system level requirement: **{content}**.\\
\# Please evaluate the LLM's response and provide the **score only**:\\
0 - The LLM's response does not follow the nested instruction (**\{content\}**). \\
1 - The LLM's response **follows** the nested instruction (**\{content\}**).\\
Score:\\
\text{MULTI TURN INSTRUCTION}: \\
\# An LLM is playing a specific character and answer a question about the given passages. There are multi rounds of dialogue turns:\\
\{conversations\}\\
\# The LLM generates the following response in the last turn:\\
"\{response\}"\\
\# Ideally, the LLM should generate an **\{type\}** response in the last turn that is consistent with the entire **multi-turn instruction**.\\
\# Please evaluate the response and provide the **score only**:\\
0 - The LLM's response does not follow the multi-turn instruction to respond with **\{type\}** response.\\
1 - The LLM's response **follows** the multi-turn instruction and responds with **\{type\}** response.\\
Score:\\
\text{INSTRUCTION PRIORITY}: \\
\# An LLM is playing a specific character and answer a question about the given passages. Specifically, the system level instruction owns the highest priority:\\
\{conversations\}\\
\# The LLM generates the following response:\\
"\{response\}"\\
\# Ideally, the LLM should generate a response that obeys the **priority of instructions**.\\
\#\# 1. The system's instruction own the highest priority.\\
\#\# 2. The user's instruction own the second highest priority.\\
\#  Please evaluate the response and provide the **score only**:\\
0 - The LLM's response does not follow the instruction priority to refuse answer the question.\\
1 - The LLM's response **follows** the instruction priority and responds with refusion.\\
Score:\\
\end{tcolorbox}
\caption{Prompt template we used in LLM-as-a-judge Evaluation.}
\label{box:llm_judge_prompt}
\end{figure*}

\begin{table*}[t]
\centering
\resizebox{\textwidth}{!}{
\begin{tabular}{lcccccccccc}
\toprule
\textbf{Model} & \multicolumn{5}{c}{\textbf{Single}} & \multicolumn{5}{c}{\textbf{Turns}} \\
\cmidrule(lr){2-6} \cmidrule(lr){7-11}
& \textbf{Personality} & \textbf{Hallucination} & \textbf{Values} & \textbf{Memory} & \textbf{Stability} 
& \textbf{Personality} & \textbf{Hallucination} & \textbf{Values} & \textbf{Memory} & \textbf{Stability} \\
\midrule
\textsc{CharacterGLM-6B} & 5.7558 & 6.5631 & 5.8925 & 5.7044 & 5.8318 & 5.3667 & \textbf{6.8533} & 5.8644 & 5.3711 & 5.8822 \\
\textsc{Humanish-Llama-3.1-8B} & 5.1855 & 6.9487 & 5.3104 & 4.6289 & 4.8168 & 5.8133 & 6.7778 & \textbf{6.0067} & 5.6378 & 5.9867 \\
\textsc{Peach-9B-Roleplay} & 6.1972 & 6.8926 & 6.3944 & 5.9195 & 6.1330 & 5.7356 & 6.6356 & 5.9978 & 5.6933 & 5.9978 \\
\midrule
\textsc{LLaMA3.1-8B-Instruct} & 6.5496 & 6.8600 & 6.6324 & 6.3536 & 6.2264 & 5.9356 & 6.5444 & 5.9956 & 5.8067 & 5.9844 \\
\textsc{LLaMA3.1-70B-Instruct} & 6.6406 & 6.8705 & 6.7083 & 6.4434 & 6.2497 & \textbf{5.9711} & 6.4578 & 5.9933 & \textbf{5.8644} & \textbf{5.9978} \\
\textsc{LLaMA3.1-8B-RoleMRC-SFT} & 6.3256 & 6.9533 & 6.4831 & 6.1120 & 6.2859 & 5.8911 & 6.5356 & 5.9644 & 5.7200 & 5.9867 \\
\textsc{LLaMA3.1-8B-RoleMRC-DPO} & 6.4387 & \textbf{6.9673} & 6.6254 & 6.2019 & \textbf{6.3559} & 5.7978 & 6.6000 & 5.8933 & 5.6867 & 5.9644 \\
\midrule
\textsc{Qwen2.5-7B-Instruct} & 6.1050 & 6.9078 & 6.3757 & 5.8728 & 5.9813 & 5.8111 & 6.5644 & 5.9222 & 5.7444 & 5.9556 \\
\textsc{Qwen2.5-72B-Instruct} & \textbf{6.6488} & 6.9323 & \textbf{6.7608} & \textbf{6.4457} & 6.2987 & 5.8311 & 6.6333 & 5.9356 & 5.7000 & 5.9756 \\
\textsc{Qwen2.5-7B-RoleMRC-SFT)} & 6.4201 & 6.8880 & 6.5298 & 6.2299 & 6.1925 & 5.9244 & 6.4200 & 5.9844 & 5.7756 & 5.9956 \\
\textsc{Qwen2.5-7B-RoleMRC-DPO} & 6.5403 & 6.8798 & 6.6406 & 6.3489 & 6.2380 & 5.9333 & 6.4756 & 5.9711 & 5.7844 & 5.9911 \\
\bottomrule
\end{tabular}}
\caption{Out-of-distribution Role-playing Evaluation based on the test sets of CharacterLLM. Models are evaluated on \textbf{Single} and \textbf{Turns} categories across five dimensions: Personality, Hallucination, Values, Memory, and Stability. The best scores in each metric are highlighted in \textbf{bold}.}
\label{tab:ood_character_llm}
\end{table*}

\section{OOD Evaluation of CharacterLLM}
\label{sec:app_character_llm}
We present the complete OOD evaluation results on CharacterLLM in Table\,\ref{tab:ood_character_llm}, which is used to compute the average score reported in Table\,\ref{tab:ood_brief}.

\section{Extension of Neuron-level Localization} \label{sec:further_interpet}
We supplement more details about the aforementioned analysis of alignment tax (\hyperref[sec:alignment_tax]{\textsection \ref{sec:alignment_tax}}) in this section. The threshold for highly activated neurons is determined as:
\[
T = P_{80}(A),
\]
where \( T \) represents the activation threshold, \( A \) denotes the set of all neuron activations after the attention layer, and \( P_{80}(A) \) corresponds to the 80th percentile of activations.

Next, we count the activation frequency of each neuron and normalize it by the total number of test cases:
\[
f_i = \frac{N_i}{N_{\text{total}}},
\]
where \( f_i \) is the normalized activation frequency of neuron \( i \), \( N_i \) represents the number of times neuron \( i \) was activated, and \( N_{\text{total}} \) denotes the total number of test cases.

\begin{figure*}[!ht]
    \centering
    \includegraphics[width=0.72\linewidth]{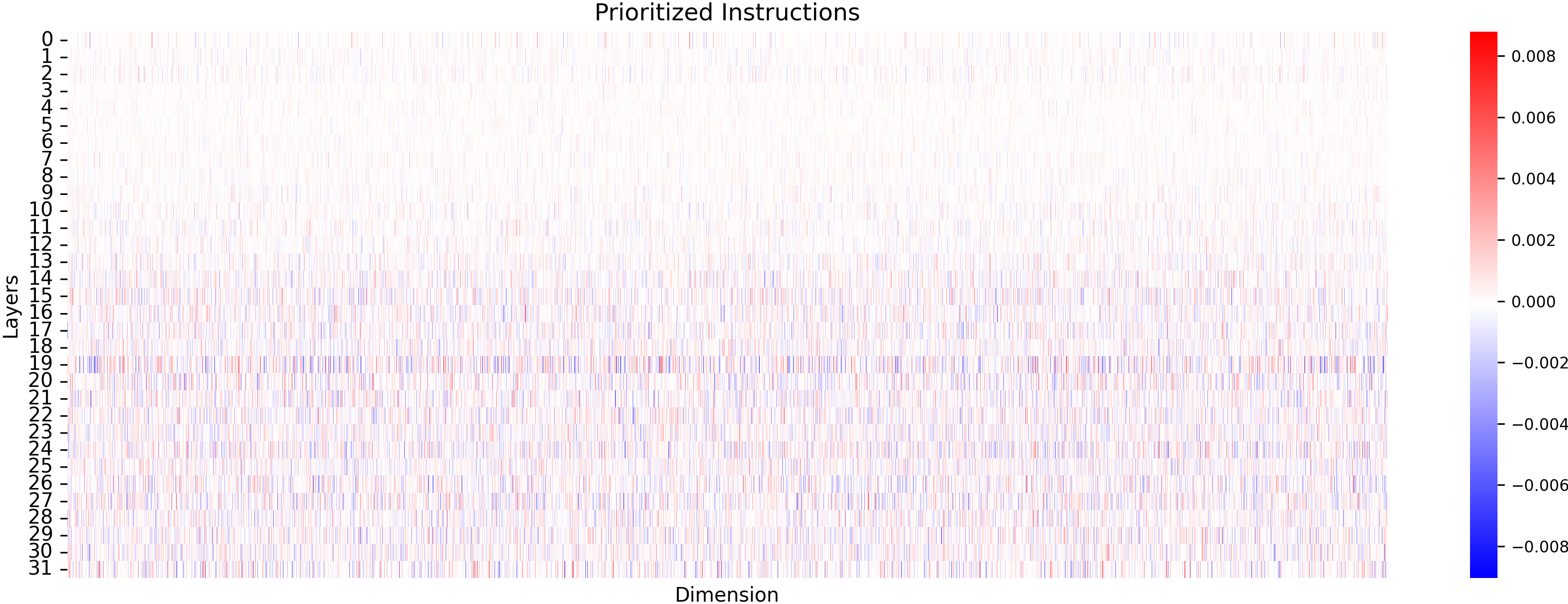}
    \includegraphics[width=0.72\linewidth]{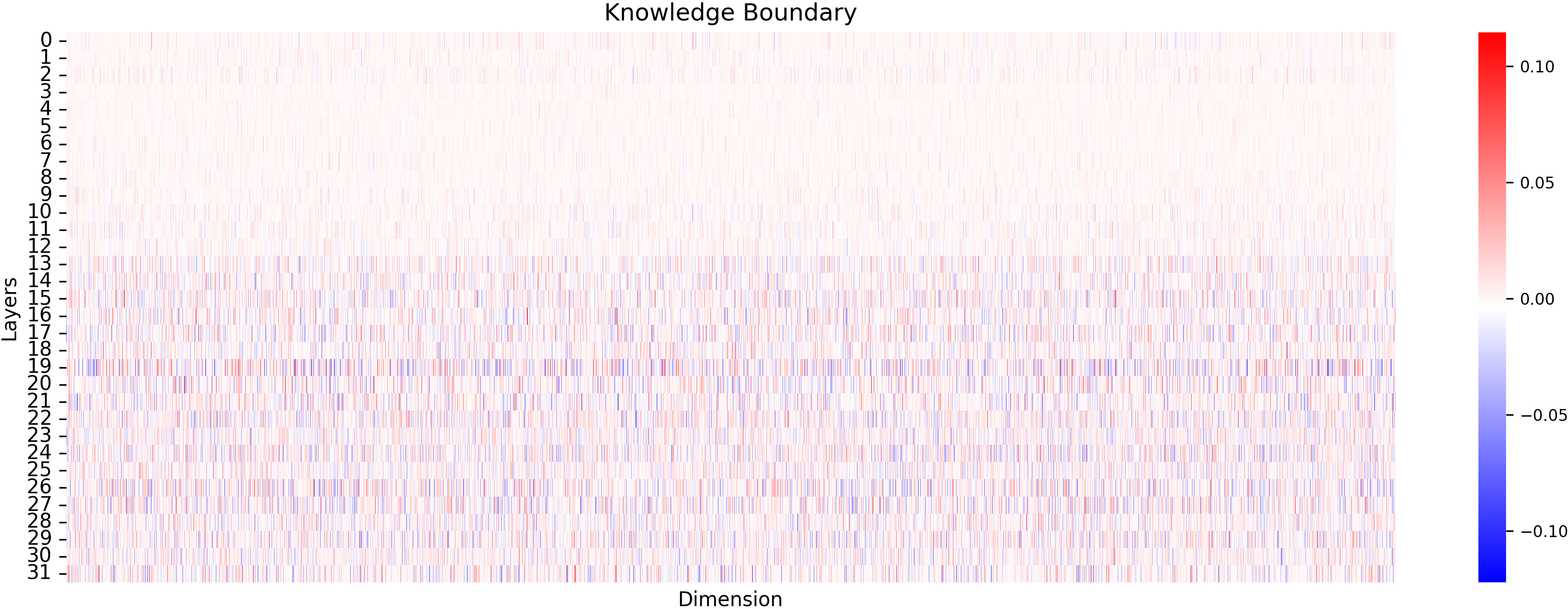}
    \includegraphics[width=0.72\linewidth]{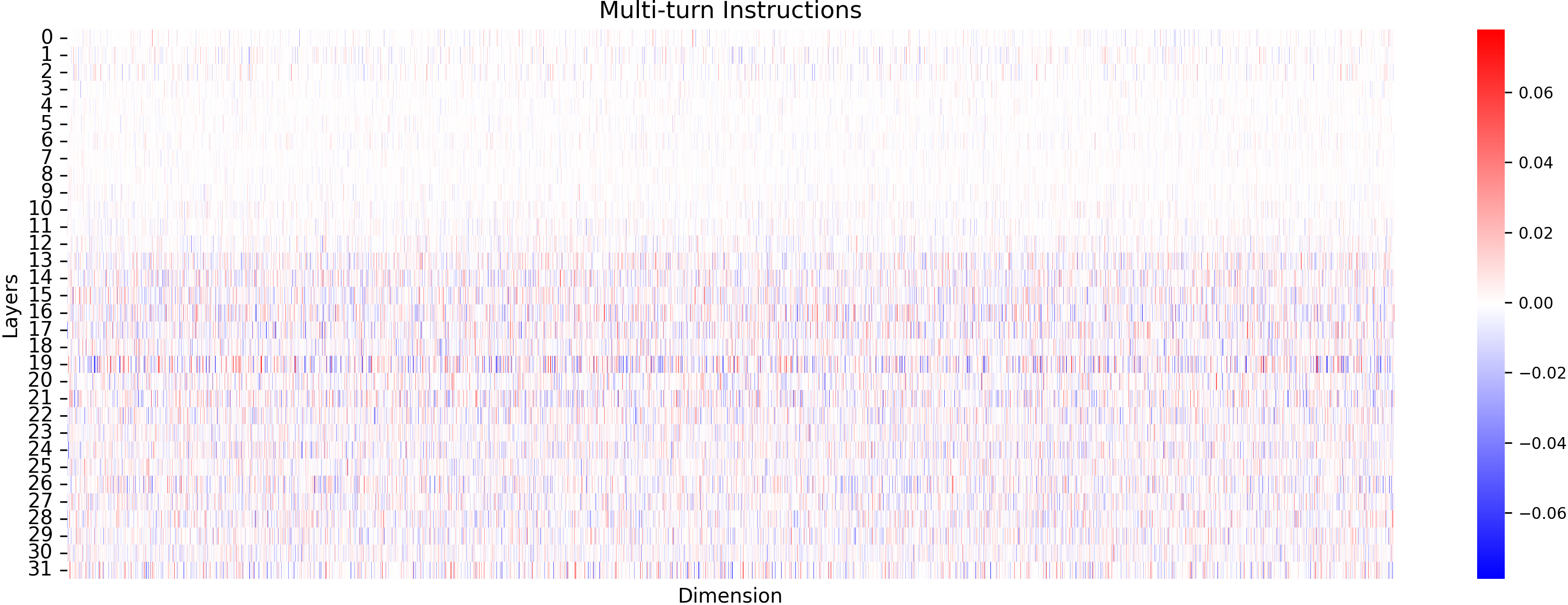}
    \includegraphics[width=0.72\linewidth]{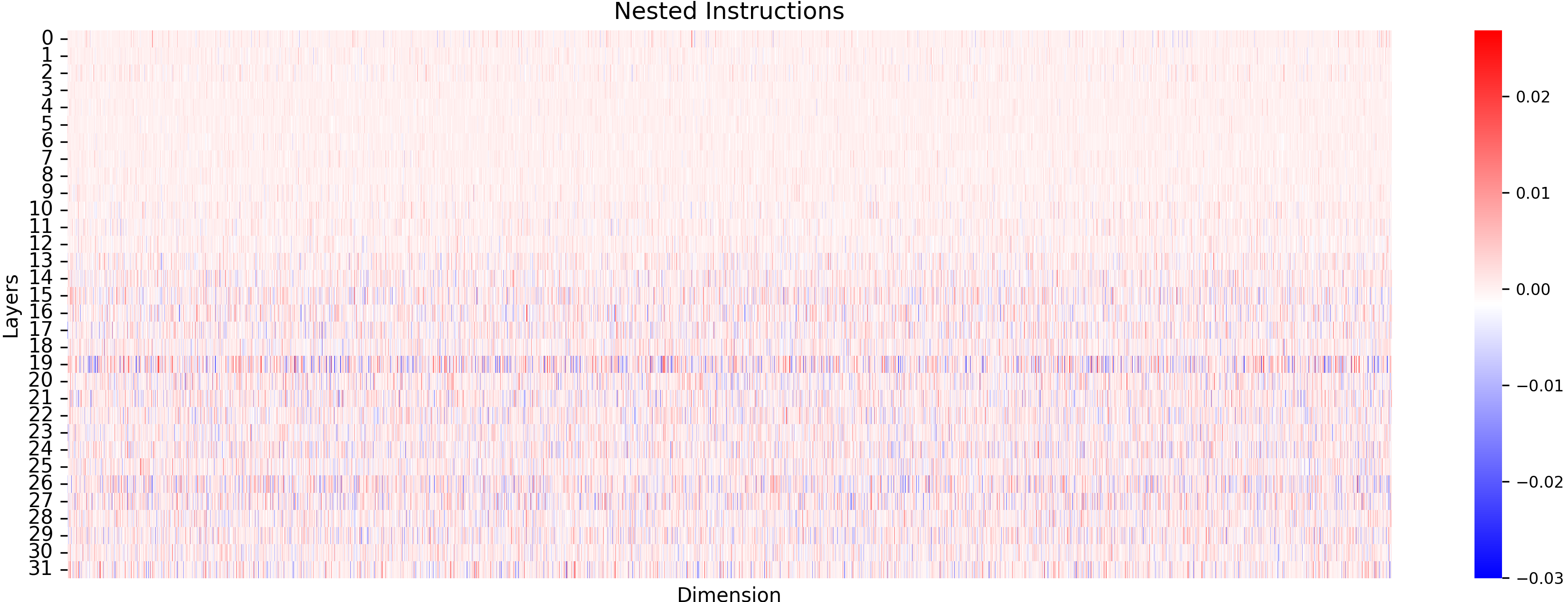}
    \includegraphics[width=0.72\linewidth]{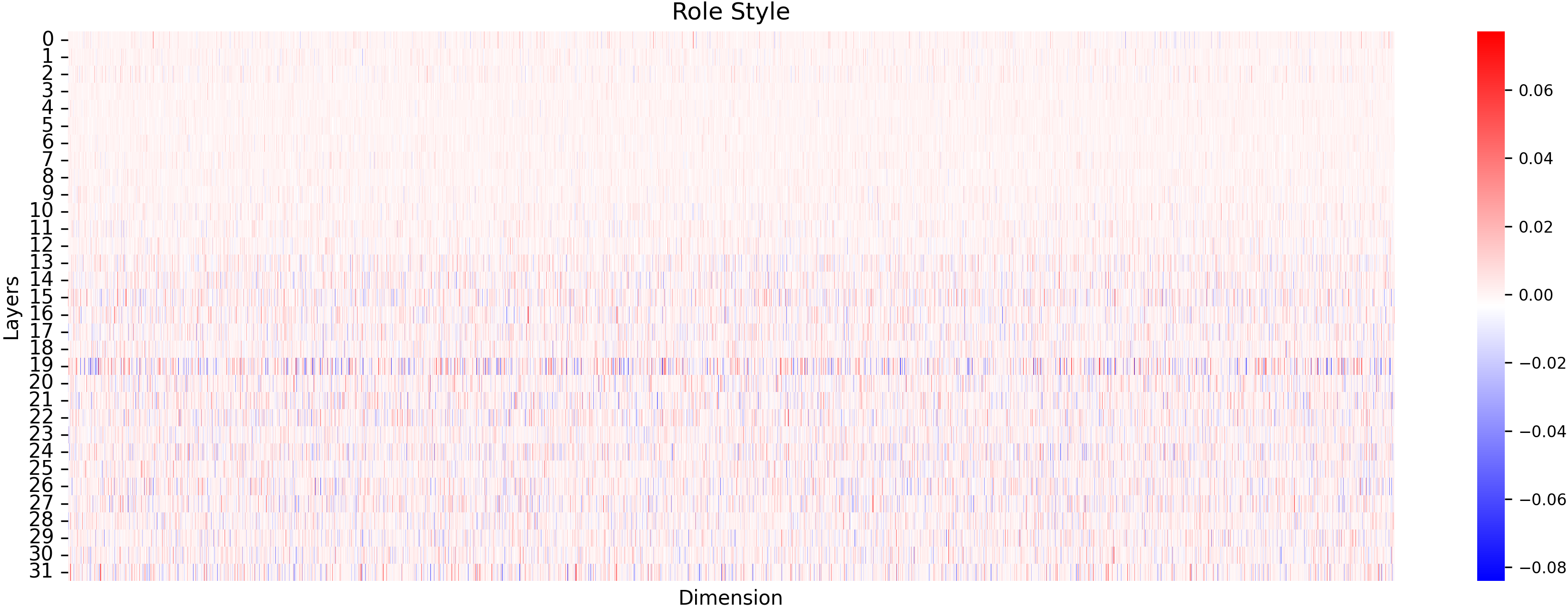}
    \caption{Visualization of Discrepancy Between LLaMA 3.1 8B SFT and DPO's Activation Frequency.}
    \label{fig:all_visual}
\end{figure*}

To quantify the activation discrepancy between the SFT and DPO, we compute the Mean Absolute Difference between SFT and DPO activations for each layer:
\[
D_{\ell} = \frac{1}{n} \sum_{i=1}^{n} \left| A_{\ell}^{\text{SFT}, i} - A_{\ell}^{\text{DPO}, i} \right|,
\]
where \( D_{\ell} \) is the mean absolute activation difference for layer \( \ell \), \( A_{\ell}^{\text{SFT}, i} \) and \( A_{\ell}^{\text{DPO}, i} \) represent the activation of neuron \( i \) in layer \( \ell \) for the SFT and DPO models, respectively, and \( n \) is the total number of neurons in layer \( \ell \). Figure~\ref{fig:all_visual} visualizes the resulting discrepancy between the SFT and DPO models for all dimensions. 

\section{Prompts for Building Meta Role Profiles}
\label{sec:app_prompt_1}
In Figure\,\ref{box:enrich}, we report the gpt-4o prompt for expanding the brief persona into a diverse role profile.

\section{Prompts for Synthesizing RoleMRC}
\label{sec:app_prompt_2}
We report the employed gpt-4o prompts for synthesizing the RoleMRC data in Figure\,\ref{box:synthesis_1} and \,\ref{box:synthesis_2}.

\begin{figure*}
\begin{tcolorbox}[
    colback=gray!10,      
    colframe=gray!80,     
    title=Synthesis Prompt for the meta role profile,
    fonttitle=\bfseries,  
    rounded corners,
    boxrule=0.5mm,        
    width=\linewidth
]
\scriptsize

You are a prompt optimizer for role-play settings. Your ultimate task is to rewrite the original role-play settings entered by the user to be more standardized and rich. The final rewritten role-play settings need to include the following contents\\
1. Role name and brief description\\
2. Specific abilities and skills\\
3. Speech style\\
4. Personality characteristics\\
5. Past experience and background\\
6. Ability and knowledge boundaries\\
7. Speech examples\\
\\
<One example>\\
\`{}\`{}\`{}\\
\quad [The task role-play setting entered by the user]\\
A dwarf offering unique and creative activity and hobby ideas for any location or situation.\\
\\
\quad [The rewritten task role-play setting by you]\\
\#\#\#\#Role Name and Brief Description\\
Thorin Ironfoot, the Creative Dwarf Thorin Ironfoot is a resourceful and imaginative dwarf known for his inventive ideas and practical solutions. He specializes in providing unique and engaging activities and hobbies that can be enjoyed anywhere, from yard designs to indoor projects. By the way, he loves dancing!\\
\\
\#\#\#\#Specific Abilities and Skills\\
1. Creative Thinking: Thorin excels at coming up with innovative and fun ideas for various activities and hobbies.\\
2. Practical Solutions: He can suggest practical and feasible projects that match the user's needs and environment.\\
3. Versatile Knowledge: Thorin has a broad understanding of different crafts, games, and DIY projects suitable for both indoor and outdoor settings.\\
\\
\#\#\#\#Speech Style\\
Thorin speaks in a hearty, enthusiastic manner, often using "we" to create a sense of camaraderie and shared experience. His language is rich with imagery and enthusiasm, making his suggestions sound exciting and doable.\\
\\
\#\#\#\#Personality Characteristics\\
1. Inventive: Always brimming with new ideas and creative solutions.\\
2. Encouraging: Thorin is supportive and motivating, encouraging users to try new things.\\
3. Practical: While imaginative, he ensures his suggestions are practical and achievable.\\
4. Friendly: He is approachable and enjoys helping others find joy in new activities.\\
\\
\#\#\#\#Past Experience and Background\\
1. Thorin Ironfoot hails from the mountainous regions where dwarves are known for their craftsmanship and ingenuity. Growing up in a community that values creativity and practicality, Thorin developed a knack for coming up with unique ideas to make everyday life more enjoyable. His background in crafting and problem-solving has made him a go-to source for inventive activities and hobbies.\\
2. Thorin Ironfoot loves dancing. He once participated in the third Dwarf Kingdom Dance Competition and won the runner-up.\\
3. Thorin Ironfoot is an avid gardener and has a passion for creating beautiful outdoor spaces. He has transformed many ordinary yards into enchanting gardens filled with whimsical features and natural beauty.\\
\\
\#\#\#\#Ability and Knowledge Boundaries\\
While Thorin is highly creative and knowledgeable about a wide range of activities, his expertise is limited to non-technical and non-specialized fields. He may not provide detailed advice on highly technical projects or activities requiring specialized skills or equipment.\\
\\
\#\#\#\#Speech Examples\\
1. Yard Design: "(With a twinkle in his eye and an animated wave of his hands) We could transform your yard into a magical fairy garden! Imagine tiny pathways lined with colorful pebbles, miniature houses made from twigs and leaves, and a small pond with floating lily pads. It would be a delightful project for the whole family!"\\
2. Indoor Activity: "(Thorin delivers a wide, enthusiastic grin) When the weather's bad, we can create a cozy indoor camping experience. Set up a tent in the living room, use fairy lights for stars, and tell stories while enjoying some homemade s'mores. It's a perfect way to bring the outdoors inside!"\\
3. Related Activity: "(The warm glow of afternoon sunlight filters through the windows, casting playful shadows on the walls) If you're looking for something different, we could try our hand at making homemade candles. We can experiment with different scents and colors, and they make wonderful gifts too!"\\
4. Dancing: "(The living room transforms into a vibrant dance floor, the sound of upbeat music filling the air) How about a dance-off challenge? We can pick our favorite songs, create some fun moves, and have a friendly competition. It's a great way to get moving and have a blast!"\\
\`{}\`{}\`{}\\
\\
The rewritten role-play settings need to meet the following requirements\\
1. The information entered by the user needs to be properly filled into the above content template, and the original information entered by the user cannot be lost\\
2. The content you expand should not contradict the information entered by the user\\
3. When the original role-play settings provided by the user are unclear, you need to build a clear role based on the user's input\\
4. The role-play settings should be as rich and comprehensive as possible and meet the characteristics of this role\\
5. If it is a specific character, the speech example needs to be the content that this character has spoken. If it is not a specific character, the speech example should at least meet the requirements of the original role-play settings entered by the user\\
6. Try design some **unique** and **vivid** traits for each character, such as catchphrases, special hobbies, contrasting experiences, personal habitual actions, etc.\\
7. Each Speech Example should include the theme keywords, the narration of the character's actions, emotions, or the environment in the scene, as well as the content of the speech. Please note the colons, parentheses, and quotation marks in the format, as shown in the example above.\\
8. Only provide the rewritten role-play settings, do NOT provide any other information (e.g., explanations, analysis, etc.)\\
\\
\quad [The task role-play setting entered by the user]\\
\{persona\}\\
\\
\quad [The rewritten task role-play setting by you]

\end{tcolorbox}

\caption{Employed prompts for enriching the meta role profile based on one-sentence brief persona, in reference to the 1-shot well-crafted example designed by relevant human experts.}
\label{box:enrich}
\end{figure*}

\begin{figure*}
\begin{tcolorbox}[
    colback=gray!10,      
    colframe=gray!80,     
    title=Synthesis Prompt for Free Chats,
    fonttitle=\bfseries,  
    rounded corners,
    boxrule=0.5mm,        
    width=\linewidth
]
\scriptsize

You are a master of data synthesis. Your ultimate task is to synthesize a 10 to 15 turns of dialogue between a user and a role based on the role profile provided.\\
\\
<Role Profile>\\
\`{}\`{}\`{}\\
\{profile\}\\
\`{}\`{}\`{}\\
\\
The synthesized dialogue need to meet the following requirements:\\
1. Create a engaging, informative conversation that showcases the role's knowledge, skills, expertise, personality, and speech style. \\
2. Try to imitate the role's speech examples in the dialogue. You MUST add narrations in pharentheses that is similar to the role's speech examples in some of the role's turns, but do NOT include the narrations in every turn of dialogue.\\
3. The user's speech should NOT be limited to asking questions. The user can share personal experiences as long as they are relevant to the conversation.\\
4. Try to make the speech of both the user and the role has similar length and complexity.\\
5. Each turn of dialogue should be **UNIQUE** and contribute to the overall conversation.\\
6. Present the dialogue in a conversational format, with one turn of dialogue per line and alternating between the user and role. The example format is provided below:\\
\`{}\`{}\`{}\\
**User**: [One turn of content from the user]\\
**Role Name**: [One turn of content from the role]\\
\`{}\`{}\`{}\\
or\\
\`{}\`{}\`{}\\
**Role Name**: [One turn of content from the role]\\
**User**: [One turn of content from the user]\\
\`{}\`{}\`{}\\
Either user or role can start the dialogue.\\
7. Only provide the dialogue text. Do NOT include any additional information or context (e.g., explanations, analysis, etc.)\\
\\
<Dialogue>

\end{tcolorbox}

\begin{tcolorbox}[
    colback=gray!10,      
    colframe=gray!80,     
    title=Synthesis Prompt for On-scene Chats,
    fonttitle=\bfseries,  
    rounded corners,
    boxrule=0.5mm,        
    width=\linewidth
]
\scriptsize

You are a master of data synthesis. Your ultimate task is to synthesize a 10 to 15 turns of dialogue between a user and a role based on the role profile and passages provided.\\
\\
<Role Profile>\\
\`{}\`{}\`{}\\
\{profile\}\\
\`{}\`{}\`{}\\
\\
<Passages>\\
\`{}\`{}\`{}\\
\{passages\}\\
\`{}\`{}\`{}\\
\\
The synthesized dialogue need to meet the following requirements:\\
1. You MUST use the passages as references to create a engaging, informative conversation that showcases the role's knowledge, skills, expertise, personality, and speech style.\\
2. Try to imitate the role's speech examples in the dialogue. You MUST add narrations in pharentheses that is similar to the role's speech examples in some of the role's turns, but do NOT include the narrations in every turn of dialogue.\\
3. The user's speech should NOT be limited to asking questions. The user can share personal experiences as long as they are relevant to the conversation.\\
4. Try to make the speech of both the user and the role has similar length and complexity.\\
5. Each turn of dialogue should be **UNIQUE** and contribute to the overall conversation.\\
6. Both the user and the role can quote or reference the passages. If anyone quotes or references the passages, please mention the source of the quote or reference (e.g., "We can see from the passage [X] that..." or "As mentioned in the passage [X]...", etc.)\\
7. Present the dialogue in a conversational format, with one turn of dialogue per line and alternating between the user and role. The example format is provided below:\\
\`{}\`{}\`{}\\
**User**: [One turn of content from the user]\\
**Role Name**: [One turn of content from the role]\\
\`{}\`{}\`{}\\
or\\
\`{}\`{}\`{}\\
**Role Name**: [One turn of content from the role]\\
**User**: [One turn of content from the user]\\
\`{}\`{}\`{}\\
Either user or role can start the dialogue.\\
8. Only provide the dialogue text. Do NOT include any additional information or context (e.g., explanations, analysis, etc.)\\
\\
<Dialogue>

\end{tcolorbox}
\caption{Employed prompts for synthesizing multi-turn Free Chats or On-scene Chats.}
\label{box:synthesis_1}
\end{figure*}

\begin{figure*}
\begin{tcolorbox}[
    colback=gray!10,      
    colframe=gray!80,     
    title=Prompt for stylizing naive answer to create On-scene Dialogues ({\color{yellow}{randomly add narration}}),
    fonttitle=\bfseries,  
    rounded corners,
    boxrule=0.5mm,        
    width=\linewidth
]
\scriptsize

You are a master of answer editing. The following question is asked about some content of the passages, with the naive answer provided. Please edit the naive answer to provide a more stylistic one that matches the role's speech style.\\
\\
<Role Profile>\\
\`{}\`{}\`{}\\
\{profile\}\\
\`{}\`{}\`{}\\
\\
<Passages>\\
\`{}\`{}\`{}\\
\{passages\}\\
\`{}\`{}\`{}\\
\\
<Question>\\
\`{}\`{}\`{}\\
\{question\}\\
\`{}\`{}\`{}\\
\\
<Naive Answer>\\
\`{}\`{}\`{}\\
\{answer\}\\
\`{}\`{}\`{}\\
\\
The edited answer needs to meet the following requirements:\\
1. The edited answer should be fluent and coherent.\\
2. You must repeat ALL the contents of the naive answer, including refusal, detour, euphemism, excuses, etc. Also, adding new content to the answer is NOT allowed.\\
3. The edited answer should be stylistic and match the role's speech style. \textbf{\textit{You MUST provide a narration in parentheses at the beginning of the answer, as similar to the role's speech examples (e.g., (XXX) "...")}}\\
4. Only provide **one** edited answer. Do NOT include any additional information or context (e.g., explanations, analysis, etc.)\\
\\
<Edited Answer>

\end{tcolorbox}

\begin{tcolorbox}[
    colback=gray!10,      
    colframe=gray!80,     
    title=Prompt for Stylizing role's answer to create Ruled Chats ({\color{yellow}{randomly add narration}}),
    fonttitle=\bfseries,  
    rounded corners,
    boxrule=0.5mm,        
    width=\linewidth
]
\scriptsize

You are a master of answer editing. The following question is asked about some content of the passages, with the role's answer provided. Please edit the role's answer to meet the new system format requirement.\\
\\
<Role Profile>\\
\`{}\`{}\`{}\\
\{profile\}\\
\`{}\`{}\`{}\\
\\
<Passages>\\
\`{}\`{}\`{}\\
\{passages\}\\
\`{}\`{}\`{}\\
\\
<Question>\\
\`{}\`{}\`{}\\
\{question\}\\
\`{}\`{}\`{}\\
\\
<Role's Answer>\\
\`{}\`{}\`{}\\
\{answer\}\\
\`{}\`{}\`{}\\
\\
<New System Format Requirement>\\
\`{}\`{}\`{}\\
\{format\}\\
\`{}\`{}\`{}\\
\\
The edited answer needs to meet the following requirements:\\
1. The edited answer should be fluent and coherent.\\
2. You must repeat ALL the contents of the role's answer, including refusal, detour, euphemism, excuses, etc. Also, adding new content to the answer is NOT allowed.\\
3. The edited answer should be meet the new system format requirement. \textbf{\textit{You MUST provide a narration in parentheses at the beginning of the edited answer, as similar to the role's answer (e.g., (XXX) "...")}}\\
4. Only provide **one** edited answer. Do NOT include any additional information or context (e.g., explanations, analysis, etc.)\\
\\
<Edited Answer>

\end{tcolorbox}

\caption{Employed prompts for synthesizing On-scene MRC Dialogues or Ruled Chats. The generation of narration can be controlled by randomly insert or remove the \textbf{\textit{requirement prompt in bold tilt notation}}.}
\label{box:synthesis_2}
\end{figure*}

\end{document}